%% file: main.tex
\documentclass{article}

\usepackage{microtype}
\usepackage{graphicx}
\usepackage{subcaption}
\usepackage{booktabs} 
\usepackage{multirow}
\usepackage{minted}
\usepackage{longtable}

\usepackage{fontawesome5}

\usepackage[table]{xcolor}
\usepackage{makecell}

\usepackage{xspace}

\def\eg{\emph{e.g.}\xspace}

\usepackage{hyperref}

\usepackage[preprint]{icml2026}

\usepackage{amsmath}

\usepackage[capitalize,noabbrev]{cleveref}

\usepackage{amssymb}
\usepackage{mathtools}
\usepackage{amsthm}

\usepackage[most]{tcolorbox} 
\usepackage{xcolor}
\definecolor{promptbg}{RGB}{245,247,250} 
\definecolor{promptframe}{RGB}{44,62,80} 
\definecolor{agentbg}{RGB}{240,244,248}
\definecolor{agentframe}{RGB}{52,73,94}

\usepackage[capitalize,noabbrev]{cleveref}

\usepackage{svg}         
\usepackage{subcaption}  

\usepackage{longtable}
\usepackage{enumitem}
\usepackage{caption} 
\usepackage{tcolorbox}
\usepackage{graphicx}  
\usepackage{amsmath}   
\usepackage{tabularx}
\usepackage{booktabs}    
\usepackage{array}       
\usepackage{ragged2e}    
\usepackage[most]{tcolorbox}
\tcbuselibrary{listings,breakable}

\usepackage{colortbl}
\usepackage[table]{xcolor}
\usepackage{array}      

\usepackage{xcolor}
\usepackage[dvipsnames]{xcolor}
\usepackage{tikz}
\usetikzlibrary{positioning,arrows.meta}

\theoremstyle{plain}

\theoremstyle{definition}

\theoremstyle{remark}

\usepackage[textsize=tiny]{todonotes}

\icmltitlerunning{Process-Supervised Multi-Agent Reinforcement Learning for Reliable Clinical Reasoning}

\begin{document}

\twocolumn[
  \icmltitle{Process-Supervised Multi-Agent Reinforcement Learning \\ for Reliable Clinical Reasoning}



  \icmlsetsymbol{equal}{*}

  \begin{icmlauthorlist}
    \icmlauthor{Chaeeun Lee}{edi}
    \icmlauthor{T. Michael Yates}{edi}
    \icmlauthor{Pasquale Minervini}{edi,miniml}
    \icmlauthor{T. Ian Simpson}{edi}
  \end{icmlauthorlist}
  \icmlaffiliation{edi}{School of Informatics, University of Edinburgh, UK}
  \icmlaffiliation{miniml}{Miniml.AI}

  \icmlcorrespondingauthor{Chaeeun Lee}{chaeeun.lee@ed.ac.uk}

  \icmlkeywords{Machine Learning, ICML}

  \vskip 0.3in
]



\printAffiliationsAndNotice{}  

\begin{abstract}
Clinical decision-making requires nuanced reasoning over heterogeneous evidence and traceable justifications. While recent LLM multi-agent systems (MAS) show promise, they largely optimise for outcome accuracy while overlooking process-grounded reasoning aligned with clinical standards. One critical real-world case of this is gene–disease validity curation, where experts must determine whether a gene is causally implicated in a disease by synthesising diverse biomedical evidence. We introduce an agent-as-tool reinforcement learning framework for this task with two objectives: (i) process-level supervision to ensure reasoning follows valid clinical pathways, and (ii) efficient coordination via a hierarchical multi-agent system. Our evaluation on the ClinGen dataset shows that with outcome-only rewards, MAS with a GRPO-trained Qwen3-4B supervisor agent substantially improves final outcome accuracy from 0.195 with a base model supervisor to 0.732, but results in poor process alignment (0.392 F1). Conversely, with process + outcome rewards, MAS with GRPO-trained supervisor achieves higher outcome accuracy (0.750) while significantly improving process fidelity to 0.520 F1. Our code is available at \url{https://github.com/chaeeunlee-io/GeneDiseaseCurationAgents}.
\end{abstract}
\begin{figure}[t]
    \centering
    \includegraphics[width=\linewidth]{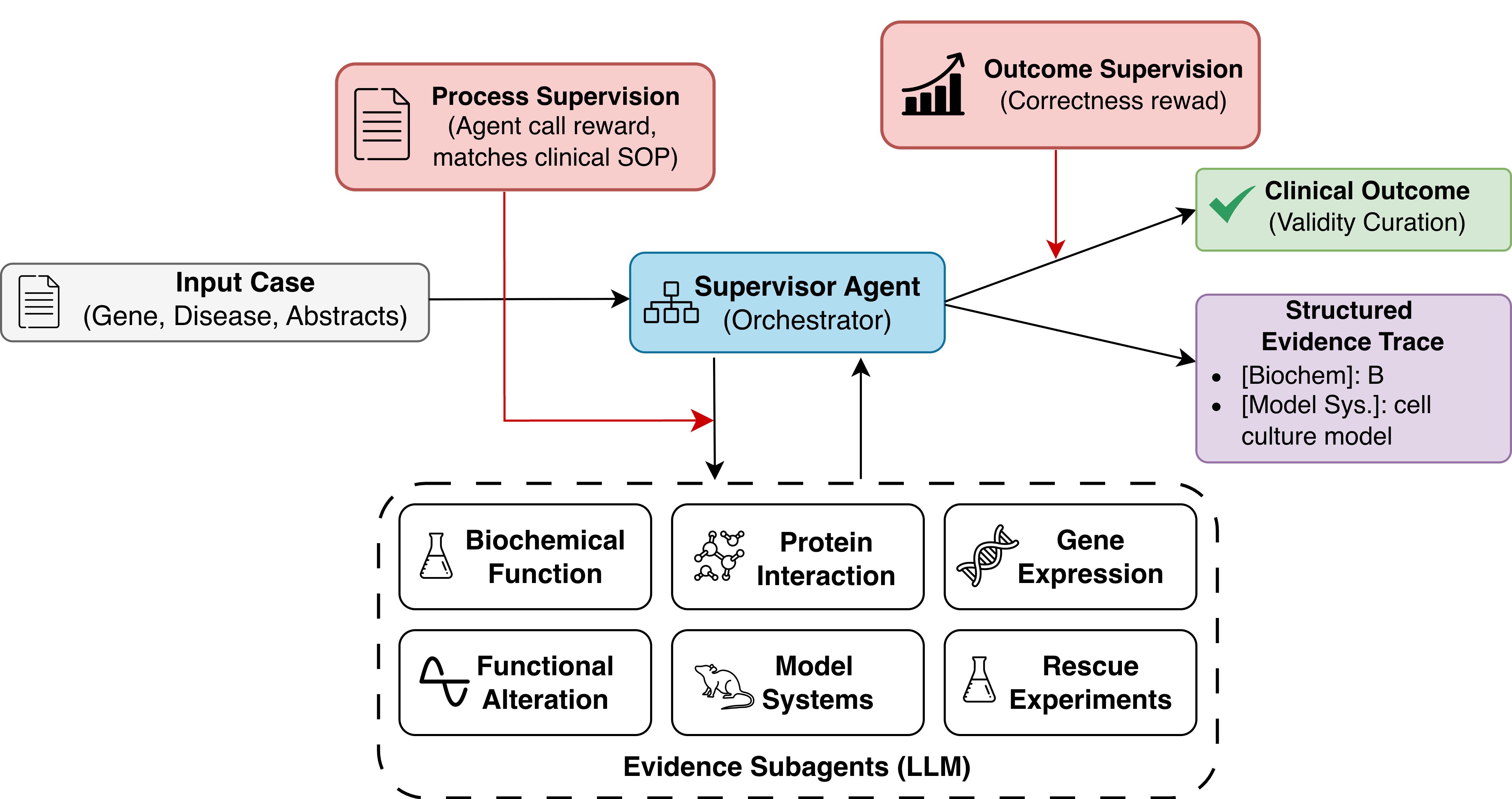}
    \caption{Overview of the hierarchical Agent-as-Tool framework for gene–disease validity curation task. The supervisor orchestrates specialised sub-agents to produce both a validity classification and a structured evidence trace matching a clinical SOP.}
    \label{fig:teaser}
\end{figure}

\section{Introduction}
Large language models (LLMs) have shown promise in biomedical applications, but their deployment in real-world clinical settings remains limited due to the need for domain expertise and the heterogeneity of biomedical reasoning tasks.
Recent works on LLM agents~\cite{yao2023reactsynergizingreasoningacting, schick2023toolformerlanguagemodelsteach, shinn2023reflexionlanguageagentsverbal, jin2025searchr1trainingllmsreason} have extended LLM reasoning through tool use, self-reflection, and retrieval, while multi-agent systems~\citep[MAS;][]{wu2023autogenenablingnextgenllm, chen2024commcollaborativemultiagentmultireasoningpath,yang2025agentnetdecentralizedevolutionarycoordination} further expand this paradigm by enabling coordination of specialised agents.
These advances provide a promising pathway for underexplored biomedical domains where process fidelity and auditability are essential \cite{hunter2016action, thaxton2022dosage, wright2024gd}.
In particular, many such settings require not only correct conclusions through complex reasoning but also structured, traceable evidence aligned with formal clinical standards.
The medical specialty of Clinical Genetics provides a representative real-world case. Rare genetic diseases (RD) affect over 300 million people worldwide \cite{irdirc2017rare}, and diagnosis of these conditions relies on establishing valid gene–disease relationships from literature.
With an increasing number of reported gene–disease relationships, developing a shared framework for consistent assessment has become increasingly important.
This need is exemplified by the Gene Curation Coalition (GenCC), an international coalition of groups engaged in gene–disease validity evaluation~\cite{gencc}, including the Clinical Genome Resource (ClinGen).
ClinGen has developed an internationally recognised standard operating procedure (SOP) that defines formal criteria for evaluating evidence supporting gene--disease relationships~\cite{clingen1, clingen2}.
These criteria decompose evidence into modular categories, providing a structured basis for decision-making.
In this work, we introduce an agent-as-tool reinforcement learning (RL) framework designed to align LLM reasoning with rigorous clinical standards.
We propose a hierarchical multi-agent architecture where a supervisor agent directs specialised sub-agents to follow valid reasoning pathways.
Each sub-agent is responsible for a distinct evidence category and produces structured outputs relevant to that category.
By training the supervisor agent with a hybrid approach that combines outcome-based and process-based rewards, we improve outcome accuracy while enforcing adherence to the clinical SOP.
In addition, the multi-agentic system produces evidence-specific reasoning traces and structured outputs that mirror clinical curation practices, providing the traceability and auditability that are crucial in real-world clinical decision-making. 

\section{Related Works}

\textbf{Biomedical Agentic Systems.}
Recent progress in LLM agents has highlighted biomedicine and healthcare as natural application areas, where many tasks benefit from agentic abilities such as literature grounding and integration with external data sources and tools.
Systems such as MMedAgent \cite{li2024mmedagentlearningusemedical}, Biomni \cite{Huang2025.05.30.656746}, AgentMD \cite{jin2024agentmdempoweringlanguageagents}, and MedAgent-Pro \cite{wang2025medagentproevidencebasedmultimodalmedical} demonstrate the potential of LLM agents to support biomedical reasoning through explicit planning, tool use, and workflow execution across a wide range of biomedical tasks \cite{rose2025meddxagentunifiedmodularagent,gao2025txagentaiagenttherapeutic}.
Studies on MAS further showcase that coordination among specialised agents can yield additional improvements, either through enhanced cognitive roles (e.g., planner, critic, verifier) or from task-specific role assignment \cite{wu2023autogenenablingnextgenllm, chen2024commcollaborativemultiagentmultireasoningpath, yang2025agentnetdecentralizedevolutionarycoordination, Liu_2025, yang2025medaideinformationfusionanatomy}. AI Hospital \cite{fan2024aihospitalbenchmarkinglarge} simulates doctor--patient interactions via a multi-agent framework to benchmark multi-turn symptom collection and diagnosis. MDAgents \cite{kim2024mdagentsadaptivecollaborationllms} introduces a multi-agent framework that assigns solo or collaborative LLM structures based on medical task complexity, improving knowledge-intensive benchmarks. Multi-agentic benchmarks such as AgentClinic \cite{schmidgall2025agentclinicmultimodalagentbenchmark} and Agent Hospital \cite{li2025agenthospitalsimulacrumhospital} facilitate medical agents under simulated environments mirroring real-world clinical scenarios. 

\textbf{Reinforcement Learning for Biomedical Agents.}
RL for LLMs has been explored as a way to further improve agentic capabilities for biomedical tasks. While studies such as SiriuS \cite{zhao2025siriusselfimprovingmultiagentsystems}, Med-R1 \cite{lai2025medr1reinforcementlearninggeneralizable}, and DoctorAgent-RL \cite{feng2025doctoragentrlmultiagentcollaborativereinforcement} report gains on task performances through RL, they largely rely on outcome-based rewards. Med-PRM \cite{yun-etal-2025-med} has shown that process-level rewards can improve performance on biomedical benchmarks by verifying intermediate steps with an LLM-as-a-judge and retrieved evidence. While DoctorAgent-RL encourages clinically structured interactions through a protocol compliance reward, it still relies on a fixed penalty structure for procedural adherence and focuses on a single agent's behavior rather than coordination of specialised agents. 
Our work incorporates process-level reward signals in a hierarchical multi-agent system using a real-world SOP-governed task to illustrate applicability in settings where process objectives are explicitly defined and established standards require modular intermediate reasoning \cite{hunter2016action, thaxton2022dosage}. While prior works such as ToolRL \cite{qian2025toolrlrewardtoollearning} have shown that tool-call rewards can improve reasoning quality in general domain, we extend this idea to settings where the tools are LLM-based agents operating within an MAS. 
\input{tables/dataset_stats}

\section{Gene--Disease Validity Assessment}
Developing AI systems that comply with established clinical standards is essential for real-world biomedical applications.
A representative example is \emph{gene--disease validity assessment}, a complex reasoning task that determines whether a candidate gene is causally implicated in a disease by synthesising evidence from the literature. Under ClinGen’s internationally recognised SOP, clinical experts derive a validity classification from a predefined outcome set $\mathcal{Y}$\footnote{We exclude categories associated with conflicting or contradictory evidence (e.g., Disputed and Refuted).} by evaluating evidence according to a consistent set of guidelines that define modular evidence categories, where
\begin{equation*}
\begin{aligned}
\mathcal{Y} = & \{\text{Definitive}, \text{Strong}, \text{Moderate}, \\ 
& \quad \text{Limited}, \text{No Known Disease Relationship}\}.
\end{aligned}
\end{equation*}
While the SOP encompasses both genetic and experimental evidence, this work focuses on experimental evidence. The detailed schema of the categories of experimental evidence defined by ClinGen's SOP is summarised in \cref{tab:experimental_categories}. 

\paragraph{Dataset Construction}
Training, development, and evaluation data are derived from the ClinGen Gene--Disease Validity Curation database \cite{clingen1, wright2024gd}.
We filtered the curated records to identify entries containing experimental evidence annotations linked to publications within the PubMed Central Open Access (PMC-OA) subset.
This process yielded a final corpus of $1{,}994$ valid articles.
We define each data instance as a case tuple $(g, d, \mathcal{D}, \mathcal{E}, y)$, comprising the unique gene--disease pair $(g, d)$, the set of associated open-access articles $\mathcal{D}$, the aggregate evidence annotations $\mathcal{E}$, and the final validity classification $y$.
ClinGen records are organised into Gene Curation Expert Panels (GCEPs), each chaired by specialists in specific clinical domains (\eg, the \emph{Hereditary Cardiovascular Disease} panel).
While all 58 active panels adhere to the SOP, the specific literature distributions and annotation nuances vary by domain. We exploit this organisational structure by assigning entire panels exclusively to the training, development, or test splits.
The held-out panel split enables evaluation of generalisation, as the model must transfer reasoning patterns to evidence curated by unseen expert groups and across novel clinical specialties.
\input{tables/experimental_categories}

\section{Process-Supervised Multi-Agent RL}
We frame the gene–disease validity curation task as a hierarchical MAS, in which a \textit{Supervisor Agent} $\pi_\theta$ orchestrates the reasoning process by invoking a set of $K=6$ specialised \textit{sub-agents} $\mathcal{A} = \{a_1, \dots, a_K\}$.
%
%
%
Each sub-agent specialises in a single evidence category (e.g., \textit{Model Systems}, \textit{Rescue}, \textit{Expression}) via prompt-based role assignment. When invoked, a sub-agent evaluates a single article for evidence of that type.
\subsection{Task Definition and Agent Trajectories} \label{sec:agent_trajectories}
An input instance is defined as a global context $x = (g, d, \mathcal{D}_{\mathrm{abs}})$, consisting of a gene $g$, a disease $d$, and a set of relevant scientific abstracts $\mathcal{D}_{\mathrm{abs}} = \{D_1, \dots, D_N\}$.
The supervisor delegates the evaluation of specific evidence categories on full-text articles to LLM sub-agents through discrete tool calls:
\begin{equation}
c = (a_k, \text{id}_j) \in \mathcal{A} \times \mathcal{D}_{\mathrm{abs}},
\end{equation}
\noindent where each call links a specific expert agent $a_k$ to a target document identifier $\text{id}_j$.
\paragraph{Supervisor Trajectory.}
Given a global input context $x = (g, d, \mathcal{D}_{\mathrm{abs}})$, the supervisor generates a trajectory:
\begin{equation}
\tau_{\text{sup}} = \Big( x,\, z_{\text{plan}},\, \mathbf{C},\, \mathbf{O},\, z_{\text{synth}},\, y \Big).
\end{equation}
Here, $z_{\text{plan}}$ is the model's reasoning before invoking the tools $\mathbf{C}$. $\mathbf{C} = \{c_1, \dots, c_m\}$ denotes a batch of parallel tool calls, while $\mathbf{O} = \{o_1, \dots, o_m\}$ represents the resulting structured \textit{observations} returned by the sub-agents. Finally, the supervisor produces $z_{\text{synth}}$ to generate the final validity classification $y \in \mathcal{Y}$. In this study, we constrain our experiments to a single round of tool usage. While the supervisor reasons in two stages (planning $z_{\text{plan}}$ and synthesis $z_{\text{synth}}$), all tool calls $\mathbf{C}$ are generated as a single parallel batch. \cref{fig:main_diagram} illustrates the supervisor trajectory and its interaction with specialised sub-agents as tool environment.
\input{figures/main_diagram}
\paragraph{Supervisor Rollout with Ground-Truth Agent Observations.}
During supervisor training, rolling out all sub-agents to obtain structured observations $\mathbf{O} = (o_1,\dots,o_m)$ can incur substantial computational overhead, particularly under Group Relative Policy Optimisation~\citep[GRPO;][]{shao2024deepseekmathpushinglimitsmathematical}, where the number of sampled trajectories scales with the group size.
To reduce this cost, we adopt a strategy in which the supervisor $\pi_\theta$ is conditioned on ground-truth sub-agent observations rather than on sampled LLM sub-agent outputs.
Formally, for a given input instance $x = (g,d,\mathcal{D}_{\mathrm{abs}})$ and a set of tool calls $\mathbf{C} = (c_1,\dots,c_m)$, where each $c_i = (a_k, \text{id}_j)$ routes agent $a_k$ to document $D_j$, we construct a synthetic observation
\begin{equation}
\tilde{\mathbf{O}} = \big(\tilde{o}_1,\dots,\tilde{o}_m\big)
\end{equation}
\noindent by retrieving from ground-truth subtype labels.
Each injected $\tilde{o}_i$ includes the correct evidence subtype label for $(a_k, D_j)$ as well as natural-language evidence summary by a clinician. This allows the supervisor to generate training time trajectories
\begin{equation}
    \tilde{\tau}_{\text{sup}} = \big(x,\, z_{\text{plan}},\, \mathbf{C},\, \tilde{\mathbf{O}},\, z_{\text{synth}},\, y \big),
    \label{eq:gt_sup_traj}
\end{equation}
\noindent reducing training costs and mitigating the noise introduced by incorrect or incomplete sub-agent predictions during training.
\paragraph{Sub-Agent Trajectory.}
During inference, each invoked sub-agent $a_k$ operates on a specific document instance with full-text visibility.
Upon receiving a call $c = (a_k, \text{id}_j)$, the sub-agent's trajectory is defined as:
\begin{equation}
\tau_{\text{sub}} = \Big( (g, d, D^{\text{full}}_j), \, z_{\text{reason}}, \, e \Big),
\end{equation}
\noindent where $D^{\text{full}}_j$ is the full text of the article specified by the supervisor in the tool argument.
Each sub-agent generates a local reasoning trace $z_{\text{reason}}$ and returns a structured evidence object $e$ which forms the observation $o \in \mathbf{O}$, where $e = (\mathcal{S},\, \mathcal{M})$.
$\mathcal{S}$ is the set of predicted evidence subtypes for document $D_j$ (possibly empty), and $M$ is an evidence-specific rationale summarising the reasoning behind the sub-agent’s conclusion, mirroring the real-world clinical curation workflow in which clinicians record natural-language summaries together with evidence annotations.
We emphasise that providing such rationales is a crucial prerequisite for real-world deployment, where evidence assessment must be transparent and incrementally updated as new studies emerge.
\input{figures/MA_reward_trajectories}

\subsection{Reward Design}
We train $\pi_\theta$ to predict the correct validity class while invoking specialised sub-agents to ground decisions in evidence categories defined by the SOP.
We optimise this with GRPO~\cite{shao2024deepseekmathpushinglimitsmathematical} using a reward $R$ that combines outcome reward with tool call-level process reward.
\paragraph{Outcome Reward ($R_{\text{out}}$).}
Outcome reward evaluates the correctness of the final validity classification $y$.
Given the ordinal nature of the ClinGen scale (Definitive $>$ Strong $>$ Moderate $>$ Limited $>$ No Known Disease Relationship), we map the label set $\mathcal{Y}$ to integer ranks $\text{rank}(\cdot) \in \{0, \dots, 4\}$ (where $0=$ \textit{No Known Disease Relationship} and $4=$ \textit{Definitive}).
We employ a centered distance-based reward:
\begin{equation}
R_{\text{out}}(y, y^*) = \sigma \cdot \left( 1 - \alpha \cdot | \text{rank}(y) - \text{rank}(y^*) | \right),
\end{equation}
\noindent where $\alpha = 0.5$ centers the base metric to $[-1, +1]$ (penalising catastrophic errors while rewarding exact matches), and $\sigma = 4.0$ scales the final reward signal.
This results in a value range of $R_{\text{out}} \in [-4, +4]$, incentivising the model to predict labels that are \textit{ordinally} close to the ground truth.
\paragraph{Process Reward ($R_{\text{proc}}$).}
The process reward $R_{\text{proc}}$ evaluates the quality of agent invocations.
Given the set of predicted tool calls $\mathbf{C}$ and ground-truth tool calls $\mathbf{C}^*$, we compute the base alignment score $s$ as the F1 score:
\begin{equation}
s = \frac{2 \cdot |\mathbf{C} \cap \mathbf{C}^*|}{|\mathbf{C}| + |\mathbf{C}^*|},
\end{equation}
\noindent where the intersection $\cap$ denotes exact matches on both agent type and input arguments.
To discourage suboptimal agent invocation strategies and accelerate convergence, we apply a cubic shaping function to the base score. Since we leverage foundational models with native tool-calling capabilities, we do not introduce complex format rewards.
Instead, we include a minimal penalty term for malformed tool blocks.
We define a tool block as the text segment enclosed within \texttt{<tool\_call>} and \texttt{</tool\_call>} tags. A block is penalised if its content fails to parse as valid JSON or if the parsed object lacks required argument keys (i.e., PMID, PMCID, gene, and disease) required by the agent schema. Denoting $n_{\text{err}}$ as the number of malformed tool blocks, the final process reward is:
\begin{equation}
R_{\text{proc}} = \text{clip}\left( \gamma \cdot s^3 - \frac{\gamma}{2} - \lambda \cdot n_{\text{err}}, -4, 4 \right),
\end{equation}
\noindent where $\gamma = 8.0$ scales the base score $s \in [0, 1]$ to $[-4, +4]$, and $\lambda = 0.5$ is the penalty coefficient for each malformed tool calls.
\paragraph{Hybrid Process + Outcome Reward ($R_{\text{Hybrid}}$).}

To incentivise the "right answer for the right reasons" trajectories, we employ a hybrid reward configuration.
The total reward $R_{\text{hybrid}}$ is computed as the weighted sum of the process and outcome components:
\begin{equation} \label{eq:rhybrid}
    R_{\text{hybrid}} = \sigma \cdot \left[ \beta R_{\text{out}} + (1-\beta) R_{\text{proc}} \right].
\end{equation}
We set the balancing coefficient $\beta=0.5$, ensuring that both the final diagnosis and the supporting evidence path contribute equally to the combined reward within the global range of $[-4, +4]$.

\subsection{Group Relative Policy Optimisation (GRPO)} \label{sec:grpo}
We optimise the supervisor policy $\pi_\theta$ using Group Relative Policy Optimisation (GRPO).
As described in \cref{sec:agent_trajectories}, the supervisor is conditioned on ground-truth observations during training. For each input $x$, the model samples a group of $G$ trajectories $\{\tilde{\tau}_1, \dots, \tilde{\tau}_G\}$ following \cref{eq:gt_sup_traj} from the policy $\pi_{\theta_{\mathrm{old}}}$.
GRPO uses the group mean reward as a baseline, removing the need for a learned value function. We exclude the KL divergence term from the loss.
The objective is:
\begin{equation}
\label{eq:grpo}
\begin{split}
\mathcal{J}_{\mathrm{GRPO}}(\theta) = & \mathbb{E}_{x \sim \mathcal{D}} \mathbb{E}_{\{\tilde{\tau}_i\}_{i=1}^G \sim \pi_{\theta_{\mathrm{old}}}} \bigg[ \frac{1}{G} \sum_{i=1}^G \min \Big( \rho_i A_i, \\
& \text{clip}(\rho_i, 1-\epsilon, 1+\epsilon) A_i \Big) \bigg],
\end{split}
\end{equation}
\noindent where the importance sampling ratio is defined over the generated tokens of the trajectory:
\begin{equation}
\rho_i = \frac{\pi_\theta(\tilde{\tau}_i \mid x)}{\pi_{\theta_{\mathrm{old}}}(\tilde{\tau}_i \mid x)}
\end{equation}
and the group-relative advantage normalises the hybrid reward $R_{\text{hybrid}}$ in \cref{eq:rhybrid} against the group statistics:
\begin{equation}
A_i = \frac{R_{\text{hybrid}}(\tilde{\tau}_i) - \mathrm{mean}(\{R_{\text{hybrid}}(\tilde{\tau}_j)\}_{j=1}^G)}{\mathrm{std}(\{R_{\text{hybrid}}(\tilde{\tau}_j)\}_{j=1}^G) + \delta}.
\end{equation}
\input{tables/main_results}
\paragraph{Single-Agent Baseline.}
We implement a single-agent pipeline as a baseline for comparison with our hierarchical multi-agent framework.
In this setting, a single agent is responsible for both identifying relevant evidence, including subtype distinctions, and producing the final validity classification $y$.
Given an input context $x = (g, d, \mathcal{D}_{\mathrm{abs}})$, the agent produces a trajectory $\tau_{\text{single}} = (x, \mathbf{C}_{\text{ft}}, \mathbf{D}_{\text{full}}, z_{\text{reason}}, J_{\text{out}})$, where $\mathbf{C}_{\text{ft}}$ denotes optional tool invocations to the function \texttt{get\_full\_text()} that retrieves the full text of an article given its identifier, and $\mathbf{D}_{\text{full}}$ denotes any retrieved full-text articles.
The response concludes with a structured JSON object $J_{\text{out}}$ that contains (i) a list of evidence items $E_{\mathrm{fine}}$ predicted at the subtype level for individual articles, and (ii) a single predicted validity label $y$.
While the process reward for the supervisor in a multi-agentic system targets high-level agent routing, the process reward in this setting directly targets fine-grained subtype predictions.
For training, we apply corresponding outcome, process, and hybrid reward formulations to $J_{\text{out}}$, with the distinction that the process component computes the base F1 score $s$ at the evidence subtype level:
\begin{equation}
s = \text{F1}(\mathbf{E}_{\text{fine}}, \mathbf{E}^*_{\text{fine}}).
\end{equation}
\cref{fig:main_diagram} illustrates the single-agent trajectory and compares it with that of the MAS supervisor.
\paragraph{Sub-Agent Supervised Fine-Tuning.} \label{sec:agent_sft}
A key difference between the single-agent and multi-agent settings lies in the granularity of the process reward.
In the single-agent setup, evidence subtypes are predicted directly, whereas the supervisor in the MAS is trained only to route queries to broader agent categories.
We additionally evaluate the effect of training both the supervisor and sub-agents.
To this end, we train a shared sub-agent LLM via supervised fine-tuning (SFT) on high-quality traces generated by DeepSeek-V3.2~\cite{deepseekai2025deepseekv32pushingfrontieropen} to fulfill all agent roles.
Although all sub-agents share the same model weights, each invocation uses a role-specific prompt and constrains the output space to the evidence subtypes relevant to that category.
\subsection{Experimental Settings}
We implement all RL and SFT experiments using the veRL framework~\citep{Sheng_2025_verl}.
To support tool use and multi-agentic topology, both single- and multi-agent settings are implemented using LangGraph, leveraging veRL's integration to enable graph-structured agentic workflows. We use Qwen3 models~\cite{yang2025qwen3technicalreport} at three scales (1.7B, 4B, and 8B) for our experiments.
All models support native tool use and strong instruction-following behaviour. Experiments with Qwen3-1.7B and Qwen3-4B were conducted on a compute node with four NVIDIA H100 (80GB) GPUs, while experiments with Qwen3-8B used a node with four NVIDIA H200 (141GB) GPUs.
\noindent \textbf{Training Configuration.}
All GRPO training utilises multi-turn rollouts with a global batch size of 16 and 8 responses per query, 5 epochs. We set the sampling temperature to 0.8 and remove KL regularisation.
For SFT, we used a micro-batch size of 1, a global batch size of 4, and a maximum sequence length of 32,600 tokens over 4 epochs. 
See Appendix~\ref{sec:training_setup} for full training details.
\paragraph{Evaluation Metrics.}
We evaluate performance at three levels of granularity: final validity classification, evidence category, and evidence subtypes.
Given a test set $\{x_i\}_{i=1}^N$, where each input is $x_i = (g_i, d_i, \mathcal{D}_i)$, \textbf{outcome accuracy} measures whether the predicted validity classification (e.g., \emph{Moderate}) matches the ground truth on a per-input basis.
\textbf{Agent call accuracy} and \textbf{evidence accuracy} are exact-set metrics averaged over inputs:
{\small
\begin{gather*}
\text{Agent Call Accuracy} = \tfrac{1}{N} \sum_{i=1}^N \mathbf{1}[C_{\text{pred}}^{(i)} = C_{\text{true}}^{(i)}] \\
\text{Evidence Accuracy} = \tfrac{1}{N} \sum_{i=1}^N \mathbf{1}[S_{\text{pred}}^{(i)} = S_{\text{true}}^{(i)}]
\end{gather*}
} \\
Here $C_{\text{pred}}^{(i)}$ and $C_{\text{true}}^{(i)}$ denote the predicted and ground-truth sets of tool calls for input $i$. Each tool call consists of a tool name (e.g., ``ExperimentalEvidence\_BiochemicalFunction\_agent'') and the required arguments (PMID, PMCID, gene, disease) supplied to the agent. $S_{\text{pred}}^{(i)}$ and $S_{\text{true}}^{(i)}$ denote the predicted and ground-truth evidence profiles for input $i$.
$S_{\text{pred}}^{(i)}$ is obtained from the outputs of the invoked sub-agents, whereas $S_{\text{true}}^{(i)}$ is defined by gold-standard annotations mapping each document in $\mathcal{D}$ to evidence subtypes (e.g., Functional Alteration \emph{Patient cells}, Biochemical Function \emph{B}).
Note that false-positive agent calls are not penalised at the evidence accuracy level if those sub-agents correctly predict no evidence.
We additionally report \textbf{macro F1} at both the agent and subtype levels, averaged across inputs, where F1 is computed over predicted versus gold agent calls and evidence subtypes, respectively.
For the \textbf{single-agent setting}, we report only \textbf{outcome accuracy} and \textbf{evidence accuracy/F1}, since the model directly predicts the final validity class and document-level subtypes.
\input{figures/outcome_vs_evidence_accuracy}
\section{Results and Analysis}
We evaluate two reward schemes (outcome-only and hybrid) across single-agent and multi-agent architectures, with results summarised in \cref{tab:main_results}.
\subsection{GRPO with Outcome-Only Reward}

In this setting, the model is trained using an outcome-based reward derived solely from the final validity class. While outcome-only training substantially improves outcome accuracy (from 0.314 to 0.786 for Qwen3-1.7B in the multi-agent setting), agents trained with outcome-only rewards exhibit poor process alignment. For example, for Qwen3-1.7B, multi-agent Agent Call accuracy drops from 0.186 in the base model to 0.114 under outcome-only GRPO, and Evidence F1 decreases from 0.353 to 0.184. Consistently across both single-agent and multi-agent configurations, outcome-only GRPO degrades process metrics (Agent Call Acc./F1, Evidence Acc./F1) relative to the untrained baseline. This suggests that outcome accuracy alone is an unreliable proxy for reasoning quality in SOP-governed tasks, where coarse outcome labels fail to capture the complexity of the task or promote adherence to the clinically defined criteria. 

\subsection{GRPO with Hybrid Reward}
GRPO with a hybrid reward combines outcome and process-level signals to encourage strong outcome accuracy while maintaining procedural validity.
We observe that this hybrid reward consistently improves process metrics while maintaining high outcome accuracy (\cref{fig:outcome_vs_evidence_accuracy}).
For example, in the multi-agent Qwen3-4B setting, Agent Call accuracy increases from 0.105 (base) and 0.082 (outcome-only) to 0.455, while Evidence F1 improves from 0.417 to 0.520, with outcome accuracy remaining stable at 0.732.
Similar trends are observed for Qwen3-8B, where Agent Call F1 improves from 0.436 (base) to 0.611 and Evidence F1 from 0.469 to 0.542. Reward trajectories also reflect that hybrid training improves both reward components (\cref{fig:reward_traj}).
\input{figures/agent_performance_breakdown}
Notably, across two agent configurations and three model sizes, training with hybrid reward improves outcome accuracy beyond outcome-only training in four of six settings.
In the multi-agent setting, as \cref{fig:agent_performance} illustrates, subtype-level Evidence Accuracy and F1 improve substantially despite the process reward being applied only at the agent-call level.
This suggests that improved high-level routing places sub-agents in more relevant contexts, indirectly enabling stronger fine-grained evidence prediction without explicit subtype-level reward.
\subsection{Sub-Agent Fine-Tuning (SFT)}
We further analyse multi-agent performance by comparing prompt-based sub-agents with sub-agents trained via SFT.
Fine-tuned sub-agents improve evidence accuracy and F1 for the 1.7B and 8B models, and increase evidence accuracy for the 4B model despite a slight decrease in F1 (\cref{tab:subagent_sft_res_main}).
These results suggest that jointly training a unified sub-agent model yields synergistic gains when combined with supervisor training.
Notably, these gains are achieved using a single shared sub-agent model with role-specific prompts, rather than training separate models for each agent role.
\input{tables/subagent_SFT_results}
\subsection{Case Studies} \label{case_studies}
To examine how the single-agent and multi-agent formulations perform under different levels of problem complexity, we present two representative case studies.
We select cases where a single article is associated with the gene–disease pair for simplicity.
Full reasoning traces from both the single-agent and multi-agent settings for selected cases are provided in the Appendix~\ref{sec:trace_examples}.

\textbf{(1) POLR1D --- Treacher Collins syndrome 2 (PMID: 27448281).}

In this example, both systems recovered the relevant \emph{Model Systems} evidence and assigned the correct validity classification (\emph{Definitive}). The evidence is clear in the abstract, well-aligned with the target disease, and does not require weighing multiple categories. In such low-complexity settings, the single-agent approach can directly determine both the outcome and the process alignment.

\textbf{(2) OCRL --- Oculocerebrorenal syndrome (PMID: 22210625).}

In contrast, this case involves multiple relevant evidence types (\emph{Model Systems} and \emph{Rescue}). While both systems still converged on the correct validity classification (\emph{Definitive}), the single-agent recovered only one of the two evidence categories. By delegating to specialised sub-agents, MAS identified an additional evidence type and produced a more complete evidence profile. This suggests that when the evidence signal is ambiguous or not directly stated in the abstract, assigning sub-agents to focus on specific evidence categories can improve process alignment.

Across both cases, the final validity classification is the same due to the strong presence of \emph{Model Systems} evidence. However, the divergence in evidence completeness in the second selected case indicates that the multi-agent approach scales more naturally to scenarios involving multiple experimental evidence types.

\section{Discussion}

\paragraph{Interpretability}
Although our evaluation focuses on outcome and process-level prediction performance, a crucial advantage of MAS lies in its ability to produce structured, evidence-specific intermediate outputs.
These intermediate rationales are indispensable for real-world clinical deployment. While a single agent can be prompted to produce both the final validity class and supporting evidence, its reasoning is not easily decomposable by evidence type, limiting interpretability at the level required for clinical curation.
By contrast, MAS yields structured outputs for each agent, including \texttt{explanation} field that mirrors how clinicians record natural-language evidence summaries.
Moreover, MAS allows reasoning traces to be inspected at the level of individual evidence categories.
This is crucial in clinical curation settings, where evidence assessments accumulate over time as new studies are published, and where intermediate evidence annotations are essential both for arriving at a final validity class and for enabling clinicians to revisit past decisions and integrate emerging evidence~\cite{clingen2}.
%



%
\section{Conclusion}
In this work, we introduce a process-supervised multi-agent RL framework for gene--disease validity curation.
Our approach trains a supervisor to coordinate specialised evidence agents using GRPO with a hybrid reward that combines outcome correctness and process-aligned agent invocation.
This hybrid approach substantially improves process fidelity while maintaining strong outcome accuracy.
We show that comparable gains can be achieved by training only the supervisor without updating sub-agent parameters, reducing the computational overhead of multi-agent learning while preserving the benefits of structured coordination.
Although a single-agent pipeline achieves higher quantitative performance, the MAS produces evidence-specific intermediate outputs that better reflect clinical curation practice.
Together, our study proposes a multi-agent solution to the gene–disease validity curation task and positions multi-agent RL as a promising direction for a broader class of SOP-governed clinical tasks \cite{hunter2016action, thaxton2022dosage} that have received limited attention in prior LLM studies.

\section*{Impact Statement}

This work aims to improve the reliability of LLM-based decision support in healthcare by aligning training objectives with the procedural requirements of a real-world clinical task. We introduce a multi-agent reinforcement learning framework for gene–disease validity curation, a real-world clinical task that underpins rare disease diagnosis and directly impacts both clinicians and patients. This task exemplifies a broader class of clinical reasoning problems that require transparent evaluation of heterogeneous evidence under strict clinical SOPs. By incorporating process-aligned rewards framed as agent invocation supervision, our approach improves process fidelity while producing structured, auditable intermediate outputs. These evidence-specific rationales are a critical requirement for real-world deployment, enabling clinicians to inspect, revise, and update decisions as new literature emerges. The system is designed to assist rather than replace clinicians' curation efforts, and was developed in close collaboration with a clinician with directly relevant domain expertise. Beyond gene–disease curation, we view this work as a first step toward tackling a broader class of SOP-governed clinical tasks, encouraging more trustworthy and clinically grounded biomedical AI systems.

\section*{Acknowledgement}
This research was funded in whole or in part by the Wellcome Trust [226083/Z/22/Z] for PARADIGM (Primary Annotated Resources to Advance Discovery In Genomic Medicine). For the purpose of open access, the author has applied a CC-BY public copyright licence to any author accepted manuscript version arising from this submission.
C.L. was supported by the United Kingdom Research and Innovation (grant EP/S02431X/1), UKRI Centre for Doctoral Training in Biomedical AI at the University of Edinburgh, School of Informatics. For the purpose of open access, the author has applied a creative commons attribution (CC BY) licence to any author accepted manuscript version arising.

\bibliography{main}
\bibliographystyle{icml2026}

\newpage
\appendix
\onecolumn



\newpage



\section{Agent Prompts}
\label{sec:prompts}
We provide role-specific prompts for each agent in both multi-agent and single-agent settings. The full input context (retrieved article abstract or full-text) and Pydantic format instruction strings for structured JSON output have been omitted for brevity.

\subsection{Supervisor Prompt}
\label{sec:supervisor_prompt}
\input{tables/suppl/supervisor_prompt}

\subsection{Sub-agent Prompt}
\label{sec:subagent_prompt}
\input{tables/suppl/subagent_prompts}

\subsection{Single-agent Baseline Prompt}
\label{sec:single_agent_prompt}
\input{tables/suppl/single_agent_prompt}
\newpage

\section{Trace Examples from Selected Case Studies}
\label{sec:trace_examples}
We present full traces from selected case studies (Section~\ref{case_studies}) to illustrate the behaviour of single-agent and multi-agent systems. All examples are generated with Qwen3-4B, GRPO-trained with hybrid reward.
\input{tables/suppl/case_study_OCRL_SA}
\input{tables/suppl/case_study_OCRL_MAS}
\input{tables/suppl/case_study_POLR1D_SA}
\input{tables/suppl/case_study_POLR1D_MAS}
\newpage

\section{Reward Trajectories}
We additionally provide reward trajectories for the single-agent setting, extended supervisor training trajectories for Qwen3-4B to illustrate convergence behaviour, and process-only reward trajectories highlighting the effect of cubic scaling relative to the original F1 score.
\label{sec:reward_trajectories}
\input{figures/suppl/SA_reward_trajectories}
\input{figures/suppl/MAS_4B_reward_trajectories_longer}
\input{figures/suppl/cubic_f1_overlay}
\newpage

\section{Training Setup}
We implement our SFT and GRPO experiments using the veRL framework. Tables \ref{tab:training_hparams} and \ref{tab:sft_training_hparams} summarise the training configurations for GRPO and sub-agent SFT, respectively. Both the single-agent GRPO and the multi-agent supervisor GRPO settings employ the same training configuration. All GRPO experiments employ multi-turn rollouts with a global batch size of 16, sampling 8 responses per query, and are trained for 5 epochs. For sub-agent SFT, we use a micro-batch size of 1, a global batch size of 4, and a maximum sequence length of 32,600 tokens over 4 epochs. The reduced batch size is necessary due to the substantially longer input sequences, as the sub-agents are prompted with the full text of the scientific article specified by the supervisor agent in the tool arguments.
\label{sec:training_setup}
\input{tables/suppl/hps}
\input{tables/suppl/SFT_hps}

\newpage

\end{document}

%% file: tables/dataset_stats.tex
\begin{table}[t]
\centering
\small
\caption{\textbf{Dataset statistics.} Breakdown of dataset construction from the ClinGen gene--disease validity curation database to the final data splits. \textit{Exp.~Ev.} = records containing experimental evidence annotations. Note that articles may overlap across splits. Only records (gene--disease pairs) with experimental evidence annotations linked to open-access article(s) are retained for the final data splits.}
\label{tab:full_dataset_stats}
\resizebox{\columnwidth}{!}{%
\begin{tabular}{l r r r r r}
\toprule
\multirow{2}{*}{\textbf{Stage}} &
\multirow{2}{*}{\textbf{\#Panels}} &
\multicolumn{2}{c}{\textbf{Records}} &
\multicolumn{2}{c}{\textbf{Unique Articles}} \\
\cmidrule(lr){3-4} \cmidrule(lr){5-6}
& & \textbf{Total} & \textbf{Exp.~Ev.} & \textbf{Total} & \textbf{PMC-OA} \\
\midrule

Raw Source & 58 & 3{,}182 & 2{,}797 & 8{,}766 & 2{,}062 \\
\cmidrule(lr){1-6}
Train & 40 & -- & 931 & -- & 1{,}345 \\
Dev   & 4  & -- & 224 & -- & 231 \\
Test  & 8  & -- & 220 & -- & 299 \\
\cmidrule(lr){1-6}
\quad Total Used & \textbf{52} & -- & \textbf{1{,}375} & -- & \textbf{1{,}994} \\
\bottomrule
\end{tabular}
}
\end{table}

%% file: tables/experimental_categories.tex
\begin{table}[t]
\small
\caption{\textbf{ClinGen's experimental evidence schema.} Experimental evidence categories and subtypes defined by ClinGen's SOP. Each category contains two or more subtype labels that may be assigned based on literature analysis.}
\label{tab:experimental_categories}
\centering
\begin{tabular}{@{}p{0.22\linewidth} p{0.60\linewidth}@{}}
\toprule
\multirow{1}{=}{\textbf{Category}} & \textbf{Subtypes} \\
\midrule

\multirow{2}{=}{Biochemical Function} &
(A) Shared biochemical function with genes implicated in the disease; \\
& (B) Biochemical function consistent with disease phenotype. \\
\midrule

\multirow{4}{=}{Protein Interaction} &
Physical association; \\
& Genetic interaction (sensu unexpected); \\
& Negative genetic interaction; \\
& Positive genetic interaction. \\
\midrule

\multirow{2}{=}{Gene Expression} &
(A) Gene expressed in tissues relevant to the disease of interest; \\
& (B) Gene altered in expression in patients who have the disease. \\
\midrule

\multirow{2}{=}{Functional Alteration} &
Patient cells; \\
& Non-patient cells. \\
\midrule

\multirow{2}{=}{Model Systems} &
Non-human model organism; \\
& Cell culture model. \\
\midrule

\multirow{4}{=}{Rescue} &
Rescue in human; \\
& Rescue in patient cells; \\
& Rescue in non-human model organism; \\
& Rescue in cell culture model. \\
\bottomrule
\end{tabular}
\end{table}

%% file: figures/main_diagram.tex
\begin{figure*}[t]
    \centering
    \includegraphics[width=0.9\textwidth]{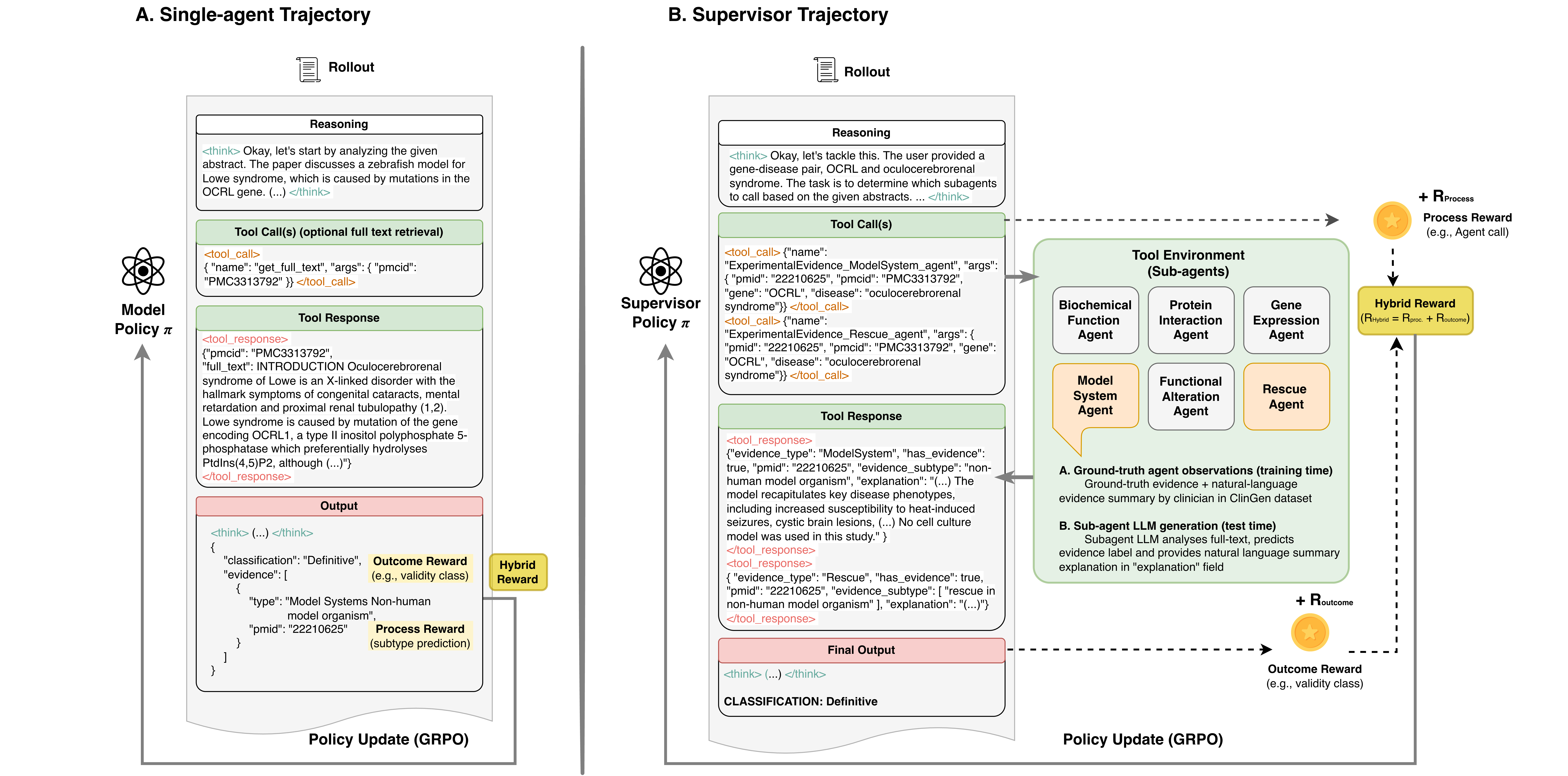}
    \caption{\textbf{Training trajectories for single-agent and multi-agent settings.}
    In the single-agent setting (left), a single model optionally retrieves full text and directly predicts evidence subtypes and the final validity class. Process reward is applied at the level of evidence subtypes. In the multi-agent setting (right), a supervisor model invokes specialised sub-agents through tool calls, and aggregates their outputs to construct (subtype-level) evidence profile and predicts the final validity class. Process reward is applied at the level of the supervisor’s agent call(s).}
    \label{fig:main_diagram}
\end{figure*}

%% file: figures/MA_reward_trajectories.tex
\begin{figure*}[t]
    \centering
    \begin{subfigure}{0.24\textwidth}
        \centering
        \includegraphics[width=\linewidth]{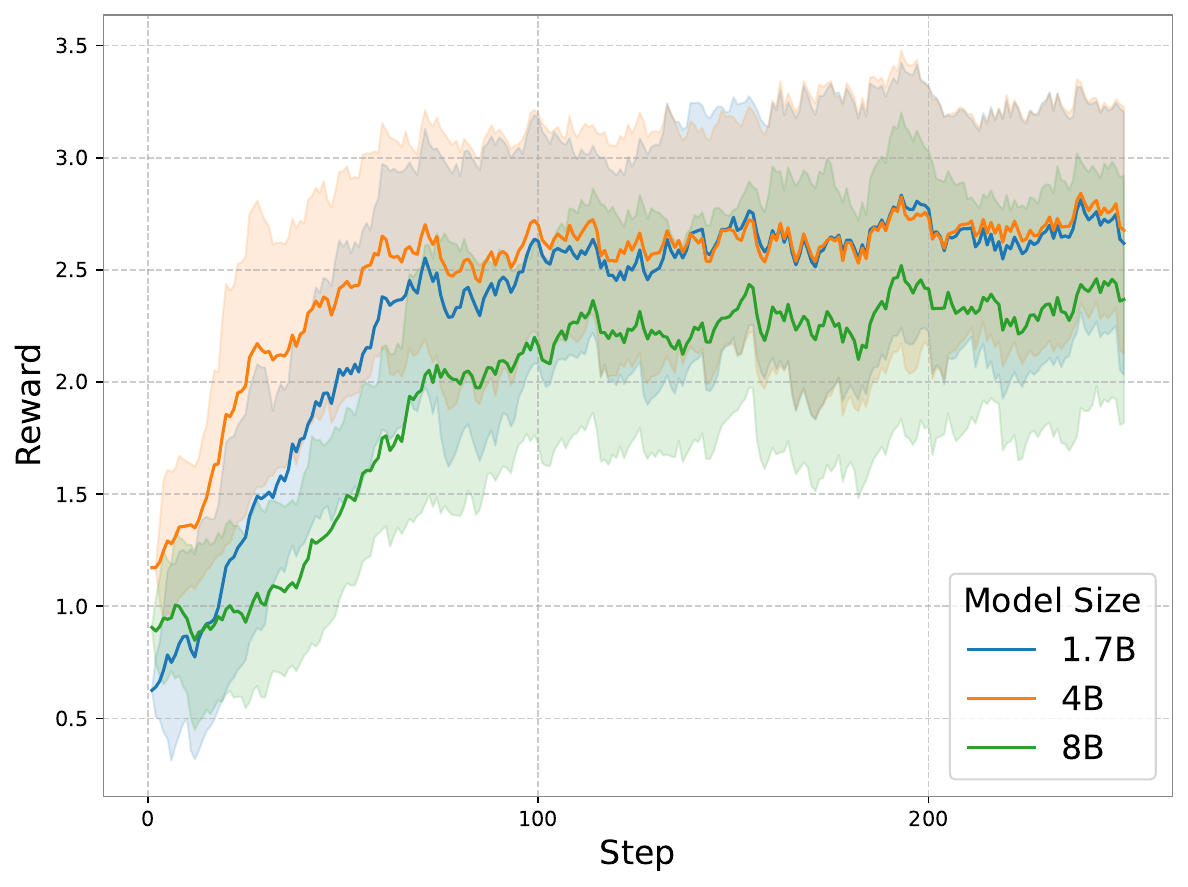}
        \caption{Outcome-Only}
    \end{subfigure}\hfill
    \begin{subfigure}{0.24\textwidth}
        \centering
        \includegraphics[width=\linewidth]{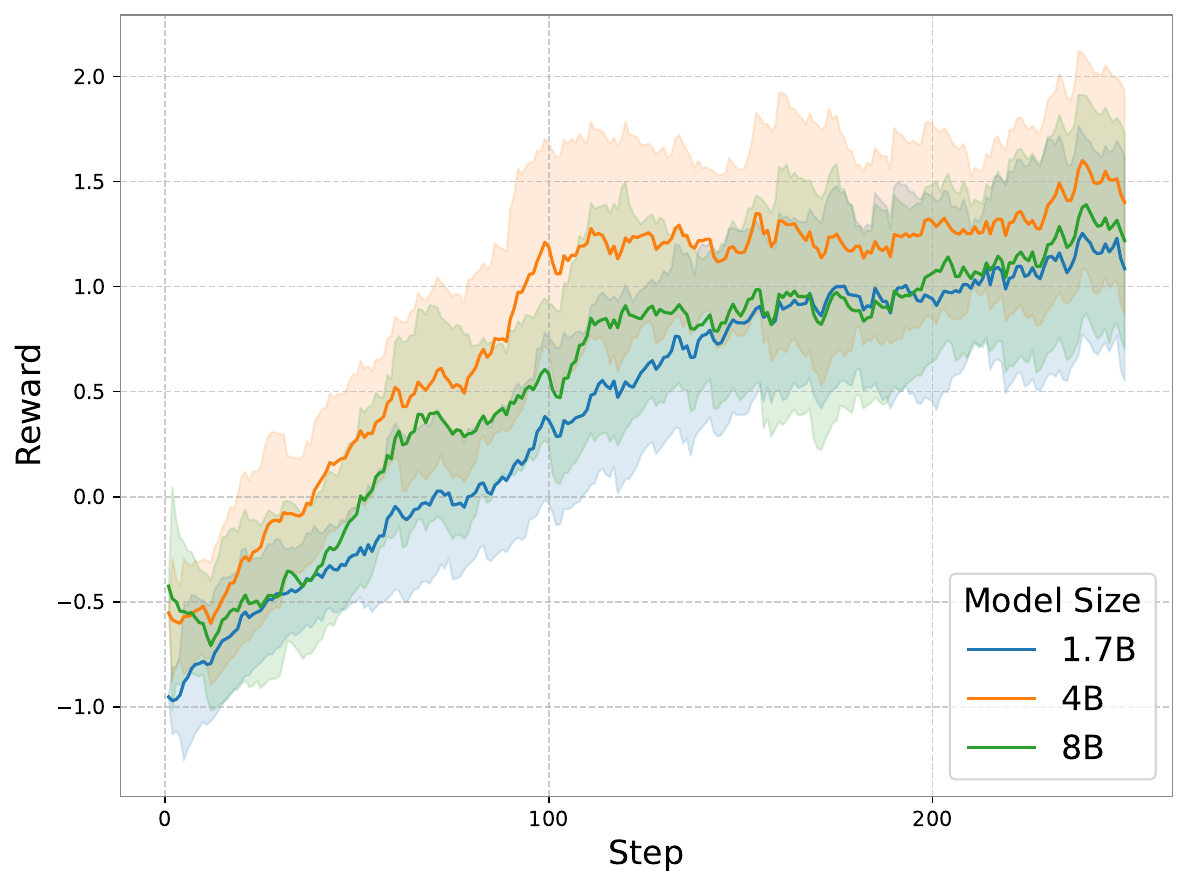}
        \caption{Hybrid (Total)}
    \end{subfigure}\hfill
    \begin{subfigure}{0.24\textwidth}
        \centering
        \includegraphics[width=\linewidth]{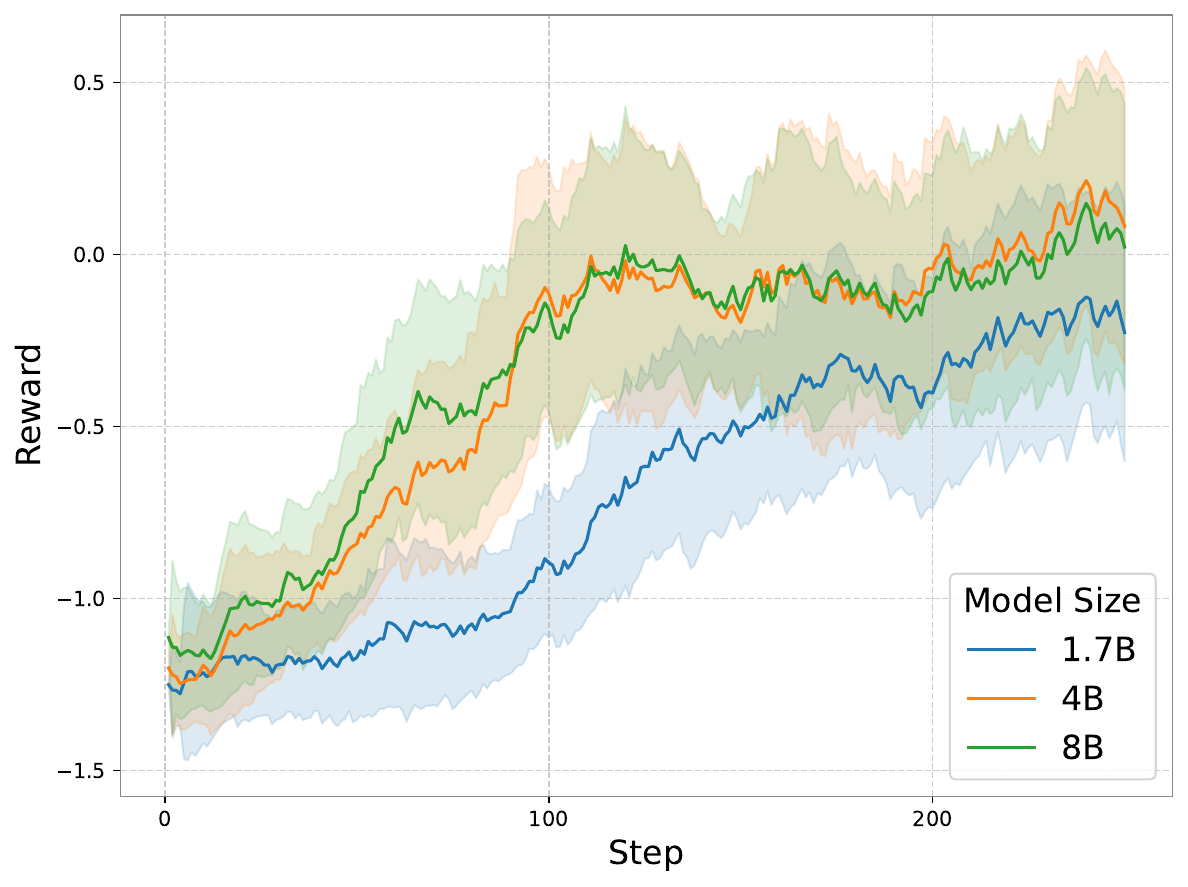}
        \caption{Hybrid (Process)}
    \end{subfigure}\hfill
    \begin{subfigure}{0.24\textwidth}
        \centering
        \includegraphics[width=\linewidth]{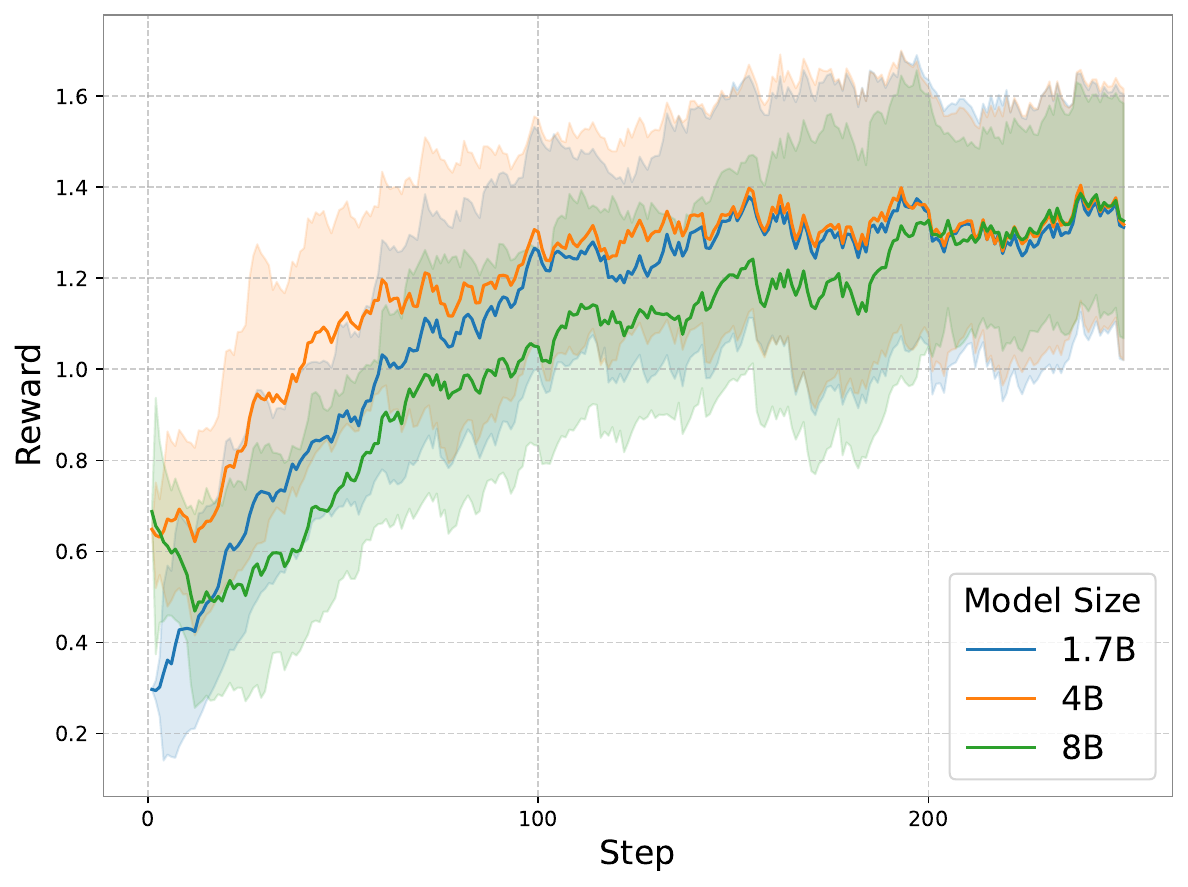}
        \caption{Hybrid (Outcome)}
    \end{subfigure}
    \caption{
        \textbf{Supervisor reward trajectories.}
        Reward progression over training steps for (a) outcome-only total reward, (b) hybrid reward, (c) process component of the hybrid reward, and (d) outcome component of the hybrid reward. 
        Longer convergence trajectories and single-agent trajectories are provided in the Appendix~\ref{sec:reward_trajectories}.
    }
    \label{fig:reward_traj}
\end{figure*}

%% file: tables/main_results.tex
\begin{table*}[t]
\centering
\caption{\textbf{Main Experimental Results.} We compare results from the proposed hybrid process + outcome reward (shaded) against the base model and an outcome-only GRPO baseline across single-agent and multi-agent settings. GRPO with hybrid reward consistently improves process metrics (Agent Call and Evidence accuracy/F1) while also improving outcome accuracy. Best metrics are \textbf{bolded}, second-best are \underline{underlined}. In all multi-agent configurations, the supervisor and sub-agents share the same base model. The supervisor model is subsequently trained in GRPO settings, while sub-agents remain frozen (base model).}

\label{tab:main_results}
\resizebox{0.98\textwidth}{!}{%
\begin{tabular}{ll|ccc|ccccc}
\toprule
\textbf{Model} & \textbf{Setting} &
\multicolumn{3}{c|}{\textbf{Single-Agent}} &
\multicolumn{5}{c}{\textbf{Multi-Agent}} \\

\cmidrule(lr){3-5} \cmidrule(lr){6-10}
& &
\makecell[c]{\textbf{Outcome}\\\textbf{Acc.}} &
\makecell[c]{\textbf{Evidence}\\\textbf{Acc.}} &
\makecell[c]{\textbf{Evidence}\\\textbf{F1}} &
\makecell[c]{\textbf{Outcome}\\\textbf{Acc.}} &
\makecell[c]{\textbf{Agent Call}\\\textbf{Acc.}} &
\makecell[c]{\textbf{Agent Call}\\\textbf{F1}} &
\makecell[c]{\textbf{Evidence}\\\textbf{Acc.}} &
\makecell[c]{\textbf{Evidence}\\\textbf{F1}} \\
\midrule

DeepSeek-V3.2 & Base Model
 & 0.209 & 0.086 & 0.451
 & 0.209 & 0.177 & 0.464 & 0.114 & 0.455 \\

GPT-5-mini & Base Model
 & 0.209 & 0.027 & 0.207
 & 0.191 & 0.114 & 0.485 & 0.095 & 0.331 \\

\midrule
\multirow{3}{*}{\textit{Qwen3-1.7B}} 
 & Base Model & 0.214 & 0.000 & 0.171 & 0.314 & 0.186 & 0.467 & 0.014 & 0.353 \\
 & GRPO (Outcome Only) & 0.777 & 0.082 & 0.164 & \textbf{0.786} & 0.114 & 0.383 & 0.109 & 0.184 \\
\rowcolor{gray!15}
 \cellcolor{white}\multirow{-1}{*}{} & \cellcolor{gray!15}GRPO (Process + Outcome)
 & \cellcolor{gray!15}0.782 & \cellcolor{gray!15}\underline{0.332} & \cellcolor{gray!15}\textbf{0.575}
 & \cellcolor{gray!15}\underline{0.750} & \cellcolor{gray!15}0.395 & \cellcolor{gray!15}0.485 & \cellcolor{gray!15}\underline{0.291}* & \cellcolor{gray!15}0.501 \\

\midrule
\multirow{3}{*}{\textit{Qwen3-4B}}
 & Base Model & 0.045 & 0.032 & 0.367 & 0.195 & 0.105 & 0.504 & 0.068 & 0.417 \\
 & GRPO (Outcome Only) & 0.782 & 0.095 & 0.286 & 0.732 & 0.082 & 0.479 & 0.045 & 0.392 \\
\rowcolor{gray!15}
 \cellcolor{white}\multirow{-1}{*}{} & \cellcolor{gray!15}GRPO (Process + Outcome)
 & \cellcolor{gray!15}\underline{0.786} & \cellcolor{gray!15}\textbf{0.382} & \cellcolor{gray!15}0.557
 & \cellcolor{gray!15}0.732 & \cellcolor{gray!15}\underline{0.455} & \cellcolor{gray!15}\underline{0.588}
 & \cellcolor{gray!15}\underline{0.291}* & \cellcolor{gray!15}\underline{0.520} \\

\midrule
\multirow{3}{*}{\textit{Qwen3-8B}}
 & Base Model & 0.077 & 0.005 & 0.324 & 0.218 & 0.132 & 0.436 & 0.095 & 0.469 \\
 & GRPO (Outcome Only) & 0.764 & 0.118 & 0.300 & 0.705 & 0.127 & 0.426 & 0.077 & 0.396 \\
\rowcolor{gray!15}
 \cellcolor{white}\multirow{-1}{*}{} & \cellcolor{gray!15}GRPO (Process + Outcome)
 & \cellcolor{gray!15}\textbf{0.795} & \cellcolor{gray!15}0.323 & \cellcolor{gray!15}\underline{0.565}
 & \cellcolor{gray!15}0.718 & \cellcolor{gray!15}\textbf{0.459} & \cellcolor{gray!15}\textbf{0.611}
 & \cellcolor{gray!15}\textbf{0.327} & \cellcolor{gray!15}\textbf{0.542} \\

\bottomrule
\end{tabular}
}
\end{table*}

%% file: figures/outcome_vs_evidence_accuracy.tex
\begin{figure}[t]
    \centering
    \includegraphics[width=\columnwidth]{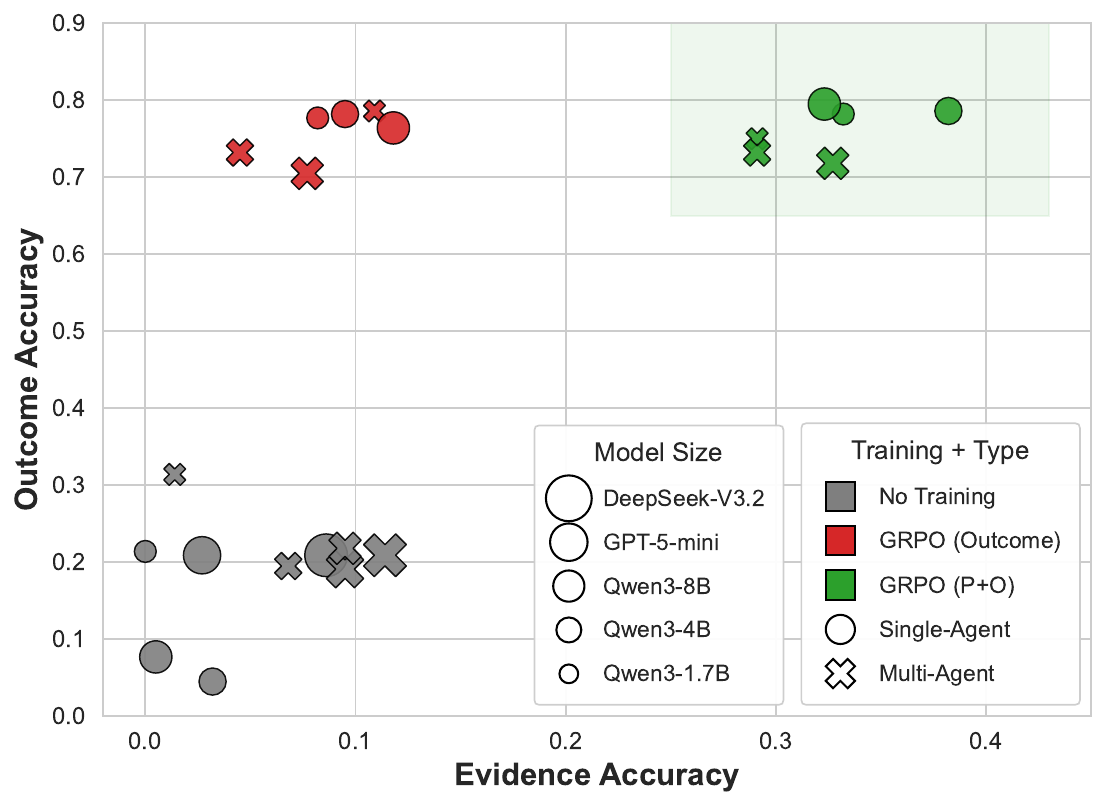}
    \caption{
        Outcome accuracy vs. evidence accuracy for single-agent and multi-agent configurations
        across model sizes and training settings. The plot illustrates how GRPO training with hybrid reward shifts performance along both axes.
    }
    \label{fig:outcome_vs_evidence_accuracy}
\end{figure}

%% file: figures/agent_performance_breakdown.tex
\begin{figure}[t]
    \centering
    \begin{subfigure}{0.48\linewidth}
        \centering
        \includegraphics[width=\linewidth]{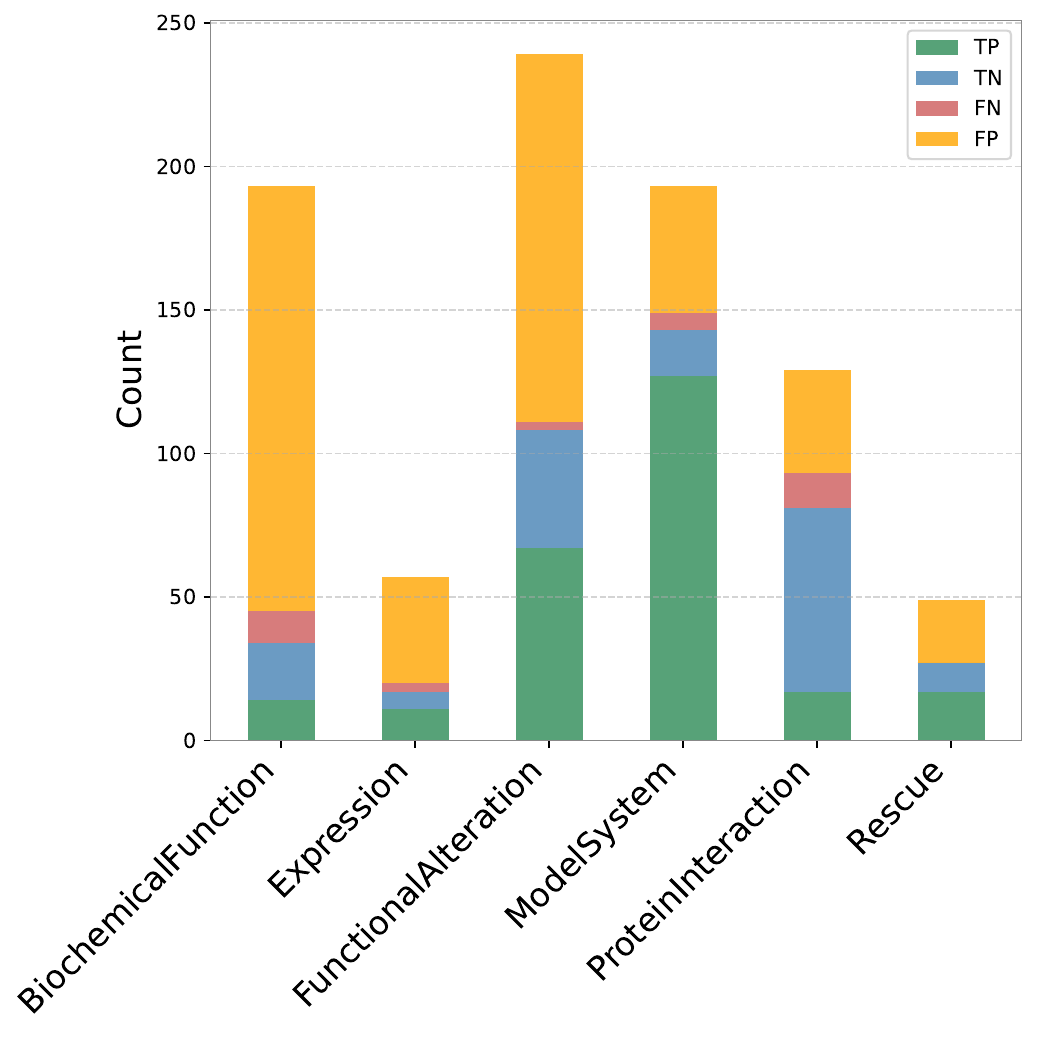}
        \caption{Outcome-Only}
    \end{subfigure}\hfill
    \begin{subfigure}{0.48\linewidth}
        \centering
        \includegraphics[width=\linewidth]{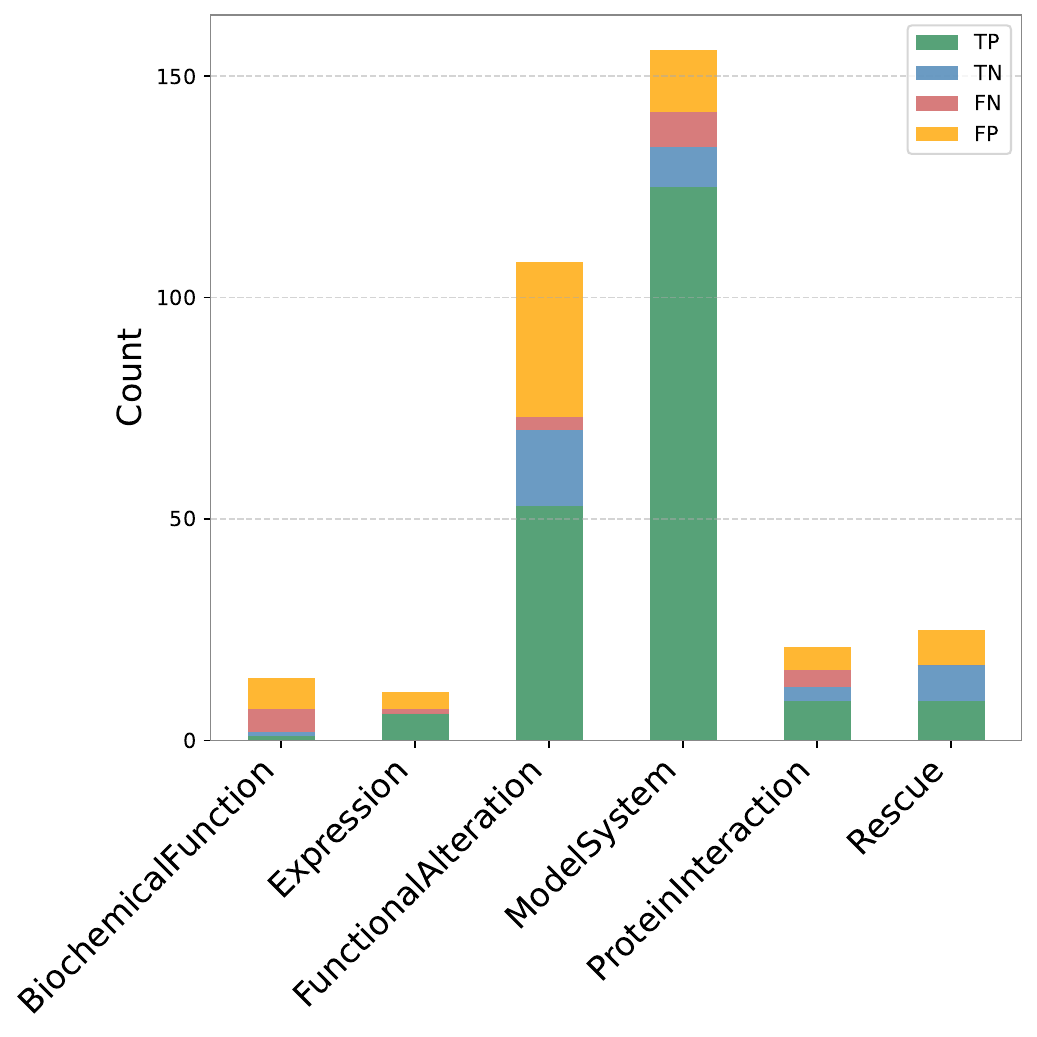}
        \caption{Process + Outcome}
    \end{subfigure}
    \caption{
        \textbf{Sub-agent performance.}
        Stacked bars show TP, TN, FN, and FP counts for each evidence agent called by supervisor trained with (a) outcome-only and (b) hybrid rewards. TP/FP/FN indicate correct/incorrect subtype predictions; TN indicates correctly predicting absence of evidence for a given category.
    }
    \label{fig:agent_performance}
\end{figure}

%% file: tables/subagent_SFT_results.tex
\begin{table}[t]
\caption{\textbf{Effect of Sub-Agent SFT} Comparison of supervisor GRPO for MAS with and without sub-agent fine-tuning. Proc+Out denotes the inference setting using a supervisor trained with hybrid process and outcome rewards, while Proc+Out+SFT further fine-tunes the sub-agent model via SFT under the same supervisor.}
\label{tab:subagent_sft_res_main}
\centering
\small
\begin{tabular}{@{}lccc@{}}
\toprule
\textbf{Model / Config} & \textbf{Outcome} & \textbf{Evidence} & \textbf{Evidence} \\
 & \textbf{Acc.} & \textbf{Acc.} & \textbf{F1} \\
\midrule
\multicolumn{4}{@{}l}{\textit{\textbf{Qwen3-1.7B}}} \\
\hspace{3mm}Proc+Out & 0.750 & 0.291 & 0.501 \\
\hspace{3mm}Proc+Out+SFT & 0.750 & \textbf{0.309} & \textbf{0.523} \\
\midrule
\multicolumn{4}{@{}l}{\textit{\textbf{Qwen3-4B}}} \\
\hspace{3mm}Proc+Out & 0.732 & 0.291 & \textbf{0.520} \\
\hspace{3mm}Proc+Out+SFT & 0.750 & \textbf{0.323} & 0.512 \\
\midrule
\multicolumn{4}{@{}l}{\textit{\textbf{Qwen3-8B}}} \\
\hspace{3mm}Proc+Out & 0.718 & 0.327 & \textbf{0.542} \\
\hspace{3mm}Proc+Out+SFT & 0.709 & \textbf{0.332} & \textbf{0.542} \\
\bottomrule
\end{tabular}
\end{table}

%% file: tables/suppl/supervisor_prompt.tex
\begingroup
\centering
\small

\label{tab:prompt_examples}

\begin{tcolorbox}[
    enhanced,
    title=\textbf{\texttt{PROMPT: Supervisor}},
    colframe=promptframe,
    colback=promptbg,
    coltitle=white,
    fonttitle=\bfseries,
    boxrule=0.5mm,
    arc=1mm,
    drop shadow,
    breakable,
    top=2mm,
    bottom=2mm,
    left=2mm,
    right=2mm
]

\ttfamily\footnotesize
\setlength{\parskip}{0.5em}

You are a supervisor agent in a gene–disease validity curation system. Your role is to evaluate the strength of monogenic gene–disease relationships using experimental evidence from the scientific literature, following clinical curation standards. \\
\\
The experimental evidences to be considered include: \\
\hspace*{2em}• Biochemical Function\\
\hspace*{2em}• Protein Interaction\\
\hspace*{2em}• Gene Expression\\
\hspace*{2em}• Functional Alteration\\
\hspace*{2em}• Model Systems\\
\hspace*{2em}• Rescue Experiments\\
As a supervisor, you can invoke subagents as tools to assist with evaluation. Each subagent is responsible for one of these six evidence categories and evaluates the corresponding evidence based on the full text of a single article.\\
\\
You will be given:\\
\hspace*{2em}• a gene–disease pair, and\\
\hspace*{2em}• a batch of one or more scientific abstracts, each prefixed with "PMID: <pmid>, PMCID: <pmcid>".\\
\\
Your task for each abstract:\\
  1. Briefly assess which experimental evidence types are likely present.\\
  2. Call only the relevant subagents for that abstract.\\
\\
Important constraints:\\
\hspace*{2em}• Call all relevant subagents immediately after assessment of all provided abstracts. There will be only one turn to call agents—you cannot make additional tool calls after this turn.\\
\hspace*{2em}• Not every abstract contains every evidence type; select only what is relevant.\\
\hspace*{2em}• Every tool call must include the correct PMID and PMCID for the abstract being evaluated.\\
\hspace*{2em}• Subagents perform detailed evaluation of the article's full text. Your role is efficient and correct routing.\\
\\
Final Classification:\\
In your second turn, after observing responses from all called subagents, provide a single final classification of the gene–disease relationship strength based on the combined experimental evidence found in full-text articles reported by the subagents.\\
\\
Choose one of:\\
\hspace*{2em}• Definitive\\
\hspace*{2em}• Strong\\
\hspace*{2em}• Moderate\\
\hspace*{2em}• Limited\\
\hspace*{2em}• No Known Disease Relationship\\
\\
End your response with the classification exactly in the following format:\\
CLASSIFICATION: <one of: Definitive, Strong, Moderate, Limited, No Known Disease Relationship>\\
\\
\end{tcolorbox}
\endgroup

%% file: tables/suppl/subagent_prompts.tex
\begin{tcolorbox}[enhanced, title=\textbf{Agent: Biochemical Function}, colframe=agentframe, colback=agentbg, drop shadow, breakable]
\ttfamily

You are an expert for evaluating the strength of evidence supporting a monogenic gene--disease relationship in literature based on biochemical function. \\
Look for any evidence showing that the gene product of \{gene\_name\} performs a biochemical function.\\
Determine whether the biochemical function is\\
\hspace*{2em}- (A) shared with other known genes in the disease of interest (\{disease\_name\}),\\
\hspace*{2em}- (B) consistent with the phenotype.\\
\hspace*{2em}- or both.\\
The biochemical function of both gene products must have been proven experimentally, and not just predicted.\\
Ensure that the other known gene(s) also have compelling evidence supporting their gene--disease relationship.\\
\end{tcolorbox}

\begin{tcolorbox}[
  enhanced,
  title=\textbf{Agent: Protein Interaction},
  colframe=agentframe,
  colback=agentbg,
  drop shadow,
  breakable
]
\ttfamily


You are an expert for evaluating the strength of evidence supporting a monogenic gene--disease relationship in literature based on protein interaction. \\
Look for any evidence showing that the gene product of \{gene\_name\} interacts with proteins previously implicated in \{disease\_name\}. \\
Proteins previously implicated in the disease of interest should have compelling evidence to support the gene-disease relationship.\\
Typical examples of this data include, but are not limited to: physical interaction via Yeast-2-Hybrid (Y2H), co-immunoprecipitation (coIP), etc.\\
NOTE: Some studies provide evidence that a variant in the gene of interest disrupts the interaction of the gene product with another protein. In these cases, the positive control, showing interaction between the two wild type proteins, can be counted as evidence of protein interaction.\\
If ProteinInteraction evidence is found, determine which interaction type(s) are reported in the article:\\
\hspace*{2em}- positive genetic interaction - An effect in which two genetic perturbations, when combined, result in a phenotype that is less severe/penetrant than would be expected from the original phenotypes, in effect making the organism more "wild type" in character with regards to the phenotype in question. \\
\hspace*{2em}- negative genetic interaction - An effect in which two genetic perturbations, when combined, result in a phenotype that is more severe/penetrant than expected given the phenotypes of the individual perturbations. \\
\hspace*{2em}- genetic interaction (sensu unexpected) - An effect in which two genetic perturbations, when combined, result in an unexpected phenotype that does not appear to be merely explained by the superimposition or addition of effects of the original perturbations.\\
\hspace*{2em}- physical association - Interaction between molecules within the same physical complex. Can be identified under conditions which suggest that the molecules are in close proximity but not necessarily in direct contact with each other.\\

\end{tcolorbox}

\begin{tcolorbox}[enhanced, title=\textbf{Agent: Gene Expression}, colframe=agentframe, colback=agentbg, drop shadow, breakable]
\ttfamily
You are an expert for evaluating the strength of evidence supporting a monogenic gene--disease relationship in literature based on gene expression. \\
Look for any evidence showing that the gene \{gene\_name\} is expressed in tissues relevant to the \{disease\_name\} and/or is altered in expression in patients who have the disease. \\
Typical examples of this data type are methods to detect RNA transcripts (RNAseq, microarrays, qPCR, qRT-PCR, Real-Time PCR), or protein expression (western blot, immunohistochemistry). \\
Determine whether\\
\hspace*{2em}- (A) the gene is expressed in tissues relevant to the disease of interest,\\
\hspace*{2em}- (B) the gene is altered in expression in patients who have the disease, \\
\hspace*{2em}- or both.\\
\end{tcolorbox}

\begin{tcolorbox}[enhanced, title=\textbf{Agent: Functional Alteration}, colframe=agentframe, colback=agentbg, drop shadow, breakable]
\ttfamily
You are an expert for evaluating the strength of evidence supporting a monogenic gene--disease relationship in literature based on functional alteration.\\
Look for any evidence showing that cultured cells, in which the function of the gene \{gene\_name\} has been disrupted, have a phenotype that is consistent with the human disease process of \{disease\_name\}. Examples include experiments involving expression of a genetic variant, gene knock-down, overexpression, etc.\\
Determine whether the functional alteration experiment was conducted in \\
\hspace*{2em}- patient cells, \\
\hspace*{2em}- non-patient cells,\\
\hspace*{2em}- or both.\\
\end{tcolorbox}

\begin{tcolorbox}[enhanced, title=\textbf{Agent: Model System}, colframe=agentframe, colback=agentbg, drop shadow, breakable]
\ttfamily
You are an expert for evaluating the strength of evidence supporting a monogenic gene--disease relationship in literature based on non-human model organism or cell culture model.\\
Look for any evidence showing that a non-human model organism or cell culture model with a disrupted copy of the gene \{gene\_name\} shows a phenotype consistent with the human disease state of \{disease\_name\}.\\
NOTE: Cell culture models should recapitulate the features of the diseased tissue e.g. engineered heart tissue, or cultured brain slices. \\
Determine whether the gene was disrupted in \\
\hspace*{2em}- a non-human model organism, \\
\hspace*{2em}- a cell culture model, \\
\hspace*{2em}- or both.\\
\end{tcolorbox}

\begin{tcolorbox}[enhanced, title=\textbf{Agent: Rescue}, colframe=agentframe, colback=agentbg, drop shadow, breakable]
\ttfamily
You are an expert for evaluating the strength of evidence supporting a monogenic gene--disease relationship in literature based on whether the relevant phenotype can be rescued.\\
Look for any evidence showing that the phenotype associated with \{disease\_name\} in humans (i.e. patients with the condition), non-human model organisms, cell culture models, or patient cells can be rescued by exogenous wild-type gene \{gene\_name\} or its gene product.\\
If the phenotype is caused by loss of function, consider evidence showing that the phenotype can be rescued by exogenous wild-type gene, gene product, or targeted gene editing. If the phenotype is caused by a gain of function variant, consider the evidence showing that a treatment which specifically blocks the action of the variant (e.g. siRNA, antibody, targeted gene editing) rescues the phenotype.\\
If Rescue evidence is found, determine which type(s) are reported in the article:\\
\hspace*{2em}- rescue in human\\
\hspace*{2em}- rescue in cell culture\\
\hspace*{2em}- rescue in non-human model organism\\
\hspace*{2em}- rescue in patient cells\\
\end{tcolorbox}

%% file: tables/suppl/single_agent_prompt.tex
\begingroup
\centering
\label{tab:prompt_single_agent}

\begin{tcolorbox}[
    enhanced,
    title=\textbf{\texttt{PROMPT: Single-Agent}},
    colframe=promptframe,
    colback=promptbg,
    coltitle=white,
    fonttitle=\bfseries,
    boxrule=0.5mm,
    arc=1mm,
    drop shadow,
    breakable,             
    top=2mm, bottom=2mm, left=2mm, right=2mm
]
\ttfamily\footnotesize
\setlength{\parskip}{0.5em}
You are an expert in a gene–disease validity curation system. Your role is to evaluate the strength of monogenic gene–disease relationship between the gene \{gene\_name\} and the disease \{disease\_name\} using experimental evidence from scientific literature, following clinical curation standards. Experimental evidence to be considered are: \\
\\
1. Biochemical Function:\\
\hspace*{2em}Evidence showing that the gene product of \{gene\_name\} performs a biochemical \hspace*{2em}function.\\
\hspace*{2em}Determine whether the biochemical function is\\
\hspace*{2em}\hspace*{2em}- (A) shared with other known genes in the disease of interest (\{disease\_name\}),\\
\hspace*{2em}\hspace*{2em}- (B) consistent with the phenotype.\\
\hspace*{2em}\hspace*{2em}- or both.\\
\hspace*{2em}The biochemical function of both gene products must have been proven \hspace*{2em}experimentally, and not just predicted.\\
\hspace*{2em}Ensure that the other known gene(s) also have compelling evidence supporting their \hspace*{2em}gene--disease relationship.\\
\\
2. Protein Interaction:\\
\hspace*{2em}Evidence showing that the gene product of \{gene\_name\} interacts with proteins \hspace*{2em}previously implicated in \{disease\_name\}. \\
\hspace*{2em}Proteins previously implicated in the disease of interest should have compelling \hspace*{2em}evidence to support the gene--disease relationship.\\
\hspace*{2em}Typical examples of this data include, but are not limited to: physical \hspace*{2em}interaction via Yeast-2-Hybrid (Y2H), co-immunoprecipitation (coIP), etc.\\
\hspace*{2em}NOTE: Some studies provide evidence that a variant in the gene of interest disrupts \hspace*{2em}the interaction of the gene product with another protein. In these cases, the \hspace*{2em}positive control, showing interaction between the two wild type proteins, can be \hspace*{2em}counted as evidence of protein interaction.\\
\hspace*{2em}If ProteinInteraction evidence is found, determine which interaction type(s) are \hspace*{2em}reported:\\
\hspace*{2em}\hspace*{2em}- Protein interactions positive genetic interaction - An effect in which \hspace*{2em}two genetic perturbations, when combined, result in a phenotype that is less \hspace*{2em}severe/penetrant than would be expected from the original phenotypes, in effect \hspace*{2em}making the organism more "wild type" in character with regards to the phenotype in \hspace*{2em}question. \\
\hspace*{2em}\hspace*{2em}- Protein interactions negative genetic interaction - An effect in which\\ \hspace*{2em}two genetic perturbations, when combined, result in a phenotype that is \\\hspace*{2em}more severe/penetrant than expected given the phenotypes of the individual \hspace*{2em}perturbations. \\
\hspace*{2em}\hspace*{2em}- Protein interactions genetic interaction (sensu unexpected) - An effect in \hspace*{2em}which two genetic perturbations, when combined, result in an unexpected phenotype \hspace*{2em}that does not appear to be merely explained by the superimposition or addition of \hspace*{2em}effects of the original perturbations.\\
\hspace*{2em}\hspace*{2em}- Protein interactions physical association - Interaction between molecules \hspace*{2em}within the same physical complex. Can be identified under conditions which suggest \hspace*{2em}that the molecules are in close proximity but not necessarily in direct contact \hspace*{2em}with each other.\\
\\
3. Expression:\\
\hspace*{2em}Evidence showing that the gene \{gene\_name\} is expressed in tissues relevant to the \hspace*{2em}\{disease\_name\} and/or is altered in expression in patients who have the disease. \\
\hspace*{2em}Typical examples of this data type are methods to detect RNA transcripts (RNAseq, \hspace*{2em}microarrays, qPCR, qRT-PCR, Real-Time PCR), or protein expression (western blot, \hspace*{2em}immunohistochemistry). \\
\hspace*{2em}Determine whether\\
\hspace*{2em}\hspace*{2em}- (A) the gene is expressed in tissues relevant to the disease of interest,\\
\hspace*{2em}\hspace*{2em}- (B) the gene is altered in expression in patients who have the disease, \\
\hspace*{2em}\hspace*{2em}- or both.\\
\\
4. Functional Alteration:\\
\hspace*{2em}Evidence showing that cultured cells, in which the function of the gene \{gene\_name\} \hspace*{2em}has been disrupted, have a phenotype that is consistent with the human disease \hspace*{2em}process of \{disease\_name\}. Examples include experiments involving expression of a \hspace*{2em}genetic variant, gene knock-down, overexpression, etc.\\
\hspace*{2em}Determine whether the functional alteration experiment was conducted in \\
\hspace*{2em}\hspace*{2em}- patient cells, \\
\hspace*{2em}\hspace*{2em}- non-patient cells,\\
\hspace*{2em}\hspace*{2em}- or both.\\
\\
5. Model System:\\
\hspace*{2em}Evidence showing that a non-human model organism or cell culture model with a \hspace*{2em}disrupted copy of the gene \{gene\_name\} shows a phenotype consistent with the human \hspace*{2em}disease state of \{disease\_name\}.\\
\hspace*{2em}NOTE: Cell culture models should recapitulate the features of the diseased tissue \hspace*{2em}e.g. engineered heart tissue, or cultured brain slices. \\
\hspace*{2em}Determine whether the gene was disrupted in \\
\hspace*{2em}\hspace*{2em}- a non-human model organism, \\
\hspace*{2em}\hspace*{2em}- a cell culture model, \\
\hspace*{2em}\hspace*{2em}- or both.\\
\\
6. Rescue:\\
\hspace*{2em}Evidence showing that the phenotype associated with \{disease\_name\} in humans \hspace*{2em}(i.e. patients with the condition), non-human model organisms, cell culture \hspace*{2em}models, or patient cells can be rescued by exogenous wild-type gene \{gene\_name\} \hspace*{2em}or its gene product.\\
\hspace*{2em}If the phenotype is caused by loss of function, consider evidence showing that \hspace*{2em}the phenotype can be rescued by exogenous wild-type gene, gene product, or targeted \hspace*{2em}gene editing. If the phenotype is caused by a gain of function variant, consider \hspace*{2em}the evidence showing that a treatment which specifically blocks the action of \hspace*{2em}the variant (e.g. siRNA, antibody, targeted gene editing) rescues the phenotype.\\
\hspace*{2em}If Rescue evidence is found, determine which type(s) are reported in the article:\\
\hspace*{2em}\hspace*{2em}- Rescue Human\\
\hspace*{2em}\hspace*{2em}- Rescue Patient cells\\
\hspace*{2em}\hspace*{2em}- Rescue Cell culture model\\
\hspace*{2em}\hspace*{2em}- Rescue Non-human model organism\\
\\
You will be given:\\
  \hspace*{2em}• a gene–disease pair, and\\
  \hspace*{2em}• a batch of one or more abstracts for scientific articles, each prefixed with "PMID: <pmid>, PMCID: <pmcid>".\\
\\
Your task for each article is to assess which experimental evidence categories and subtypes are present. You can invoke given tool `get\_full\_text` if additional detail is required for the article.\\
\\
Important constraints:\\
\hspace*{2em}• After assessing all provided abstracts, call `get\_full\_text` for every article that requires full-text review. There will be only one turn to call tools—you cannot make additional tool calls after this turn.\\
\hspace*{2em}• Every tool call must include the correct PMCID for the article. \\
\\
Final Classification:\\
In your second turn, based on the assessed evidence, provide a single final classification of the gene–disease relationship strength.\\
\\
Choose one of:\\
  \hspace*{2em}• Definitive\\
  \hspace*{2em}• Strong\\
  \hspace*{2em}• Moderate\\
  \hspace*{2em}• Limited\\
  \hspace*{2em}• No Known Disease Relationship\\
\\
End your response by summarising the identified evidence types and the final classification formatted as a JSON instance like the following:\\
\{\\
\hspace*{2em}"classification": "<one of: Definitive, Strong, Moderate, Limited, No Known Disease Relationship>",\\
\hspace*{2em}"evidence": [\\
\hspace*{2em}\hspace*{2em}\{\\
\hspace*{2em}\hspace*{2em}\hspace*{2em}"type": "<evidence\_type\_1>",\\
\hspace*{2em}\hspace*{2em}\hspace*{2em}"pmid": "<source\_pmid>"\\
\hspace*{2em}\hspace*{2em}\},\\
\hspace*{2em}\hspace*{2em}\{\\
\hspace*{2em}\hspace*{2em}\hspace*{2em}"type": "<evidence\_type\_2>",\\
\hspace*{2em}\hspace*{2em}\hspace*{2em}"pmid": "<source\_pmid>"\\
\hspace*{2em}\hspace*{2em}\}\\
\hspace*{2em}]\\
\}\\
\\
The "evidence" list may contain any number of entries. Each entry must be an object containing:\\
\hspace*{2em}1. "type": A string that exactly matches one of the allowed evidence types below.\\
\hspace*{2em}2. "pmid": The PMID of the specific article where this evidence was found.\\
\\
Allowed evidence types:\\
\texttt{[}"Biochemical Function A", "Biochemical Function B", "Protein interactions genetic interaction (sensu unexpected)", "Protein interactions negative genetic interaction", "Protein interactions physical association", "Protein interactions positive genetic interaction", "Expression A", "Expression B", "Functional Alteration Non-patient cells", "Functional Alteration Patient cells", "Model Systems Cell culture model", "Model Systems Non-human model organism", "Rescue Patient cells", "Rescue Cell culture model", "Rescue Human", "Rescue Non-human model organism"\texttt{]}
\end{tcolorbox}
\endgroup

%% file: tables/suppl/case_study_OCRL_SA.tex
\begin{longtable}{p{0.95\textwidth}}
\caption{Single-Agent Trace Example for OCRL--oculocerebrorenal syndrome}\\
\toprule
\multicolumn{1}{c}{\textbf{\large Trace Example}} \\
\midrule

\textbf{\textsc{Input Context}} \\
\textbf{Gene:} OCRL \hfill \textbf{PMID:} 22210625 \\
\textbf{Disease:} oculocerebrorenal syndrome \\
\midrule

\textbf{\textsc{Model Reasoning}} ($z_{\mathrm{reason}}$) \\
\textless think\textgreater

Okay, let's start by analyzing the given abstract. The paper discusses a zebrafish model for Lowe syndrome, which is caused by mutations in the OCRL gene. The study shows that when OCRL1 is suppressed in zebrafish embryos, they exhibit neurological defects similar to those in Lowe syndrome patients. They also mention that OCRL1 deficiency leads to reduced Akt signaling, increased apoptosis, and decreased proliferation in neural tissues. The results suggest that OCRL1's catalytic activity and clathrin binding are essential for its function in these processes.

Now, looking at the allowed evidence types, the key here is identifying which category this falls into. The study describes a model organism (zebrafish) where OCRL1 is disrupted, leading to a disease-like phenotype. This aligns with the "Model Systems" category, specifically "Model Systems Non-human model organism" since zebrafish is a non-human model. The paper also mentions rescue experiments, but the main finding is the phenotypic consequence of OCRL1 disruption in a non-human model. Therefore, this would fall under "Model Systems Non-human model organism" evidence. 

The abstract does not mention rescue experiments, but rather the phenotypic outcome in a model organism. Therefore, the evidence type is "Model Systems Non-human model organism". The PMID is 22210625, and the PMCID is PMC3313792. So, this abstract provides evidence under the "Model Systems" category, specifically "Non-human model organism".\\
\textless /think\textgreater\\

\midrule
\textbf{\textsc{Model Output}} ($e$) \\
\begin{minipage}{\linewidth}
\footnotesize\ttfamily
\{\\
\hspace*{1em}"classification": "Definitive",\\
\hspace*{1em}"evidence": [\\
\hspace*{2em}\{\\
\hspace*{3em}"type": "Model Systems Non-human model organism",\\
\hspace*{3em}"pmid": "22210625"\\
\hspace*{2em}\}\\
\hspace*{1em}]\\
\}
\end{minipage}\\

\midrule

\textbf{\textsc{Evaluation}} \\[0.25em]

\textbf{Validity Class:} Definitive (pred) vs Definitive (GT)%
\ \textcolor{ForestGreen}{\faCheck} \\[0.5em]

\textbf{Evidence Comparison (PMID 22210625)}\\
\textbf{Ground-truth evidence:} 
[Model Systems -- non-human model organism;\ Rescue -- non-human model organism] \\
\textbf{Predicted evidence:} 
[Model Systems -- non-human model organism] \\
\textbf{Missed evidence:} 
[Rescue -- non-human model organism] \\[0.25em]
\bottomrule
\end{longtable}

%% file: tables/suppl/case_study_OCRL_MAS.tex
\begin{longtable}{p{0.95\textwidth}}
\caption{Multi-Agent Trace Example for OCRL--oculocerebrorenal syndrome.}\\
\toprule
\multicolumn{1}{c}{\textbf{\large Trace Example}} \\
\midrule

\textbf{\textsc{Input Context}} \\
\textbf{Gene:} OCRL \hfill \textbf{PMID:} 22210625 \\
\textbf{Disease:} oculocerebrorenal syndrome \\
\midrule

\textbf{\textsc{Supervisor Planning}} ($z_{\mathrm{plan}}$) \\
\textless think\textgreater

Okay, let's tackle this. The user provided a gene-disease pair, OCRL and oculocerebrorenal syndrome. The task is to determine which subagents to call based on the given abstracts. 

First, I need to look at the abstract provided. The abstract is from a 2012 paper in Hum Mol Genet. The title mentions a zebrafish model for Lowe syndrome. The abstract discusses how OCRL1 deficiency in zebrafish embryos leads to neurological defects similar to Lowe syndrome patients. It also mentions Akt signaling, apoptosis, and proliferation issues in neural tissue. Rescue experiments are mentioned, showing that catalytic activity and clathrin binding are essential.

Now, looking at the available subagents. The ModelSystem agent is for non-human models with disrupted genes showing phenotypes consistent with the disease. The Rescue agent is for experiments where phenotypes are rescued. The abstract talks about a zebrafish model (ModelSystem) and rescue experiments. So I need to call both ModelSystem and Rescue agents for this abstract.

Wait, but the user said to call only relevant subagents. The ModelSystem agent is for model systems with disrupted genes. The Rescue agent is for rescue experiments. The abstract mentions both. So that's two tool calls here. Let me check the parameters: each tool call needs the correct PMID and PMCID. The abstract's PMID is 22210625 and PMCID is PMC3313792. So both ModelSystem and Rescue agents would be called with these values.

I should make sure there are no other subagents relevant. The abstract doesn't mention biochemical functions, protein interactions, gene expression, or functional alterations. So only ModelSystem and Rescue are appropriate here. Therefore, two tool calls for this abstract.\\
\textless /think\textgreater\\

\midrule

\textbf{\textsc{Agent Calls}} ($\mathbf{C}$) \\

\vspace{0.1em} \\

\begin{minipage}[t]{0.48\linewidth}
\{\\[0pt]
\hangindent=1.5em \hangafter=1 "name": "ExperimentalEvidence\_ModelSystem\\\_agent",\\[0pt]
\hangindent=1.5em \hangafter=1 "args": \{\\[0pt]
\hangindent=3em \hangafter=1 "pmid": "22210625",\\[0pt]
\hangindent=3em \hangafter=1 "pmcid": "PMC3313792",\\[0pt]
\hangindent=3em \hangafter=1 "gene": "OCRL",\\[0pt]
\hangindent=3em \hangafter=1 "disease": "oculocerebrorenal syndrome"\\[0pt]
\hangindent=1.5em \hangafter=1 \},\\[0pt]
\hangindent=1.5em \hangafter=1 "type": "tool\_call"\\[0pt]
\}
\end{minipage}
\hfill
\begin{minipage}[t]{0.48\linewidth}
\{\\[0pt]
\hangindent=1.5em \hangafter=1 "name": "ExperimentalEvidence\_Rescue\_agent",\\[0pt]
\hangindent=1.5em \hangafter=1 "args": \{\\[0pt]
\hangindent=3em \hangafter=1 "pmid": "22210625",\\[0pt]
\hangindent=3em \hangafter=1 "pmcid": "PMC3313792",\\[0pt]
\hangindent=3em \hangafter=1 "gene": "OCRL",\\[0pt]
\hangindent=3em \hangafter=1 "disease": "oculocerebrorenal syndrome"\\[0pt]
\hangindent=1.5em \hangafter=1 \},\\[0pt]
\hangindent=1.5em \hangafter=1 "type": "tool\_call"\\[0pt]
\}
\end{minipage}\\
\midrule

\textbf{\textsc{Subagent Observations}} ($\mathbf{O}$) \\
\begin{minipage}{\linewidth}
\{\\[0pt]
\hangindent=1.5em \hangafter=1 "evidence\_type": "ModelSystem",\\[0pt]
\hangindent=1.5em \hangafter=1 "has\_evidence": true,\\[0pt]
\hangindent=1.5em \hangafter=1 "pmid": "22210625",\\[0pt]
\hangindent=1.5em \hangafter=1 "evidence\_subtype": "non-human model organism",\\[0pt]
\hangindent=1.5em \hangafter=1 "explanation": "The article describes a zebrafish model for Lowe syndrome (oculocerebrorenal syndrome) by disrupting the OCRL1 gene. The model recapitulates key disease phenotypes, including increased susceptibility to heat-induced seizures, cystic brain lesions, reduced brain and eye size during embryogenesis, impaired neural development (reduced neural tissue cross-sectional area, increased apoptosis, reduced proliferation), and defects in Akt signaling. Rescue experiments show that catalytic activity and clathrin binding are required for OCRL1 function in these processes. No cell culture model was used in this study."\\[0pt]
\}

\vspace{0.5em}

\{\\[0pt]
\hangindent=1.5em \hangafter=1 "evidence\_type": "Rescue",\\[0pt]
\hangindent=1.5em \hangafter=1 "has\_evidence": true,\\[0pt]
\hangindent=1.5em \hangafter=1 "pmid": "22210625",\\[0pt]
\hangindent=1.5em \hangafter=1 "evidence\_subtype": ["rescue in non-human model organism"],\\[0pt]
\hangindent=1.5em \hangafter=1 "explanation": "The article reports rescue experiments in a zebrafish model of Lowe syndrome (oculocerebrorenal syndrome). Specifically, re-expression of wild-type GFP-tagged zebrafish OCRL1 rescues the morphological defects (reduction in brain size, eyes, and neural tissue) observed in OCRL1-deficient zebrafish embryos. The rescue is at least partially effective, demonstrating that exogenous wild-type OCRL can compensate for loss-of-function phenotypes in a non-human model organism."\\[0pt]
\}
\end{minipage}\\
\midrule

\textbf{\textsc{Synthesis}} ($z_{\mathrm{synth}}$) \\
\textless think\textgreater
Okay, let's see. The user provided a gene-disease pair, OCRL and oculocerebrorenal syndrome. They also gave an abstract from a study. My job is to evaluate the strength of the gene-disease relationship based on the experimental evidence mentioned.

Looking at the abstract, the study uses a zebrafish model to mimic Lowe syndrome. The ModelSystem subagent would assess this, checking if the model accurately recapitulates the disease phenotypes. The Rescue subagent would look into whether reintroducing the gene rescues the phenotypes. 

The responses from the subagents confirm that the zebrafish model shows key defects like seizures, brain lesions, and neural issues. The rescue experiments show that wild-type OCRL can fix these defects. This strong evidence from both model systems and rescue experiments suggests a strong link between OCRL and the disease.

Since both ModelSystem and Rescue evidence are present and robust, the classification should be "Definitive." There's no mention of other evidence types like biochemical functions or protein interactions, but the available evidence is sufficient and comprehensive. So the final classification is Definitive.
\textless /think\textgreater
\\
\midrule

\textbf{\textsc{Validity Class}} ($y$) \\
CLASSIFICATION: Definitive \\
\midrule

\textbf{\textsc{Evaluation}} \\[0.25em]

\textbf{Validity Class:} Definitive (pred) vs Definitive (GT)%
\ \textcolor{ForestGreen}{\faCheck} \\[0.5em]

\textbf{Evidence Comparison (PMID 22210625)}\\
\textbf{Ground-truth evidence:} 
[Model Systems -- non-human model organism;\ Rescue -- non-human model organism] \\
\textbf{Predicted evidence:} 
[Model Systems -- non-human model organism;\ Rescue -- non-human model organism] \\
\bottomrule
\end{longtable}

%% file: tables/suppl/case_study_POLR1D_SA.tex
\begin{longtable}{p{0.95\textwidth}}
\caption{Single-Agent Trace Example for POLR1D--Treacher Collins syndrome 2}\\
\toprule
\multicolumn{1}{c}{\textbf{\large Trace Example}} \\
\midrule

\textbf{\textsc{Input Context}} \\
\textbf{Gene:} POLR1D \hfill \textbf{PMID:} 27448281 \\
\textbf{Disease:} Treacher Collins syndrome 2 \\
\midrule

\textbf{\textsc{Model Reasoning}} ($z_{\mathrm{reason}}$) \\
\textless think\textgreater

Okay, let me start by analyzing the abstract provided. The study discusses the roles of Polr1c and Polr1d in craniofacial development and zebrafish models of Treacher Collins syndrome. The key findings include that homozygous mutants of polr1c and polr1d in zebrafish exhibit craniofacial anomalies, which are characteristic of the human disease. The authors also mention that the loss of function leads to defects in ribosome biogenesis and neural crest cell migration. 

The abstract mentions that the zebrafish models are used to study the disease. This suggests that the study provides evidence from a model system, which would fall under the "Model Systems" category. Specifically, the article is showing that a non-human model (zebrafish) with a disrupted copy of the gene (polr1d) exhibits a phenotype consistent with the human disease. That aligns with the "Model Systems Non-human model organism" evidence type. 

The article also describes that the loss of function in the zebrafish model leads to craniofacial defects, which are similar to those in humans. This would support the "Model Systems" category, as the study is using a non-human model to show the disease phenotype. Therefore, the evidence type here is "Model Systems Non-human model organism". 

I need to check if the abstract mentions any other types of evidence, like rescue or functional alteration. The abstract describes the mutants and their phenotypes, which is more about the model system's outcome rather than rescue. However, the study is establishing the zebrafish as a model, which is a key part of the "Model Systems" category. 

So, the primary evidence type here is "Model Systems Non-human model organism". The article provides a model system where the gene is disrupted in a non-human organism (zebrafish), and the resulting phenotype is consistent with the human disease. This fits under the "Model Systems" category, specifically the "Non-human model organism" type. 

I should make sure there's no mention of rescue or other types. The abstract doesn't mention rescue, but rather the establishment of a model system. Therefore, the evidence type is "Model Systems Non-human model organism". 

The PMID is 27448281, and the PMCID is PMC4957770. So, the evidence from this article is "Model Systems Non-human model organism".\\
\textless /think\textgreater\\

\midrule
\textbf{\textsc{Model Output}} ($e$) \\
\begin{minipage}{\linewidth}
\footnotesize\ttfamily
\{\\
\hspace*{1em}"classification": "Definitive",\\
\hspace*{1em}"evidence": [\\
\hspace*{2em}\{\\
\hspace*{3em}"type": "Model Systems Non-human model organism",\\
\hspace*{3em}"pmid": "27448281"\\
\hspace*{2em}\}\\
\hspace*{1em}]\\
\}
\end{minipage}\\

\midrule

\textbf{\textsc{Evaluation}} \\[0.25em]

\textbf{Validity Class:} Definitive (pred) vs Definitive (GT)%
\ \textcolor{ForestGreen}{\faCheck} \\[0.5em]

\textbf{Evidence Comparison (PMID 27448281)}\\
\textbf{Ground-truth evidence:} 
[Model Systems -- non-human model organism] \\
\textbf{Predicted evidence:} 
[Model Systems -- non-human model organism] \\[0.25em]
\bottomrule
\end{longtable}

%% file: tables/suppl/case_study_POLR1D_MAS.tex
\begin{longtable}{p{0.95\textwidth}}
\caption{Multi-Agent Trace Example for POLR1D--Treacher Collins syndrome 2}\\
\toprule
\multicolumn{1}{c}{\textbf{\large Trace Example}} \\
\midrule

\textbf{\textsc{Input Context}} \\
\textbf{Gene:} POLR1D \hfill \textbf{PMID:} 27448281 \\
\textbf{Disease:} Treacher Collins syndrome 2 \\
\midrule

\textbf{\textsc{Supervisor Planning}} ($z_{\mathrm{plan}}$) \\
\textless think\textgreater

Okay, let's see. The user provided a gene-disease pair, POLR1D and Treacher Collins syndrome 2. They also gave an abstract with a PMID and PMCID. My task is to determine which subagent to call for this abstract.

First, I need to look at the abstract's title and content. The title mentions "The Roles of RNA Polymerase I and III Subunits Polr1c and Polr1d in Craniofacial Development and in Zebrafish Models of Treacher Collins Syndrome." The abstract talks about zebrafish mutants with polr1c and polr1d loss-of-function leading to craniofacial defects. The study also mentions that these mutants have cartilage hypoplasia and cranioskeletal anomalies similar to humans with Treacher Collins syndrome. 

Looking at the available subagents, the ModelSystem agent is for non-human models with disrupted genes showing phenotypes consistent with the disease. The abstract discusses zebrafish models, so that's a strong candidate. The study's findings about the mutant zebrafish phenotypes matching the disease in humans fit the ModelSystem category. 

I should check if there's any other possible subagent. The Rescue agent would be if the phenotype is rescued, but the abstract doesn't mention that. The FunctionalAlteration agent is for cultured cells with gene function disrupted, but here it's zebrafish. So ModelSystem is the right choice. 

I need to make sure I'm not missing any other possible agents. The abstract also mentions ribosome biogenesis and Tp53-dependent cell death, but those are more about the mechanism. The key point is the zebrafish model's phenotype matching the disease. So the ModelSystem agent is definitely the right one here. 

So the tool call should be to ExperimentalEvidence\_ModelSystem\_agent with the given PMID and PMCID.
\textless /think\textgreater\\

\midrule

\textbf{\textsc{Agent Calls}} ($\mathbf{C}$) \\

\vspace{0.1em} \\

\begin{minipage}[t]{0.8\linewidth}
\{\\[0pt]
\hangindent=1.5em \hangafter=1 "name": "ExperimentalEvidence\_ModelSystem\_agent",\\[0pt]
\hangindent=1.5em \hangafter=1 "args": \{\\[0pt]
\hangindent=3em \hangafter=1 "pmid": "27448281",\\[0pt]
\hangindent=3em \hangafter=1 "pmcid": "PMC4957770",\\[0pt]
\hangindent=3em \hangafter=1 "gene": "POLR1D",\\[0pt]
\hangindent=3em \hangafter=1 "disease": "Treacher Collins syndrome 2"\\[0pt]
\hangindent=1.5em \hangafter=1 \},\\[0pt]
\hangindent=1.5em \hangafter=1 "type": "tool\_call"\\[0pt]
\}
\end{minipage}\\
\midrule

\textbf{\textsc{Subagent Observations}} ($\mathbf{O}$) \\
\begin{minipage}[t]{1\linewidth}
\{\\[0pt]
\hangindent=1.5em \hangafter=1 "evidence\_type": "ModelSystem",\\[0pt]
\hangindent=1.5em \hangafter=1 "has\_evidence": true,\\[0pt]
\hangindent=1.5em \hangafter=1 "pmid": "27448281",\\[0pt]
\hangindent=1.5em \hangafter=1 "evidence\_subtype": "non-human model organism",\\[0pt]
\hangindent=1.5em \hangafter=1 "explanation": "The study describes the generation and characterization of polr1d homozygous mutant zebrafish (polr1dhi2393Tg). These mutants exhibit craniofacial anomalies (smaller eyes, hypoplastic jaws, microphthalmia) and hypoplastic craniofacial cartilages, which are consistent with the craniofacial defects observed in Treacher Collins syndrome 2. The authors also show that polr1d loss-of-function leads to increased Tp53-dependent neuroepithelial cell death and deficiency of migrating neural crest cells, underpinning the cranioskeletal defects. These phenotypes are recapitulated in zebrafish and are consistent with the known pathogenesis of TCS2."\\[0pt]
\}
\end{minipage}\\
\midrule

\textbf{\textsc{Synthesis}} ($z_{\mathrm{synth}}$) \\
\textless think\textgreater
Okay, let's see. The user provided a gene-disease pair: POLR1D and Treacher Collins syndrome 2. They also gave an abstract from a study that's relevant. The task is to evaluate the strength of the gene-disease relationship based on experimental evidence.

First, I need to check the evidence types mentioned in the abstract. The study uses zebrafish models with polr1d mutants, which show craniofacial defects similar to TCS2. That's a model system. The subagent for Model Systems was called and confirmed the evidence. 

Are there other evidence types? The abstract mentions biochemical functions like ribosome biogenesis and interactions with Tp53. But the user's subagents are for specific categories. Since the main evidence here is the model system, and the other categories aren't addressed in the provided abstract, I should focus on that.

The classification needs to be based on the combined evidence. Since the study provides a non-human model with clear phenotypes matching the disease, the evidence is strong. There's no conflicting data here. So the classification should be "Definitive" because the model system directly links POLR1D to TCS2 with clear phenotypes and mechanisms.
\textless /think\textgreater
\\
\midrule

\textbf{\textsc{Validity Class}} ($y$) \\
CLASSIFICATION: Definitive \\
\midrule

\textbf{\textsc{Evaluation}} \\[0.25em]

\textbf{Validity Class:} Definitive (pred) vs Definitive (GT)%
\ \textcolor{ForestGreen}{\faCheck} \\[0.5em]

\textbf{Evidence Comparison (PMID 27448281)}\\
\textbf{Ground-truth evidence:} 
[Model Systems -- non-human model organism] \\
\textbf{Predicted evidence:} 
[Model Systems -- non-human model organism] \\
\bottomrule
\end{longtable}

%% file: figures/suppl/SA_reward_trajectories.tex
\begin{figure*}[!htb]
    \centering
    \begin{subfigure}{0.24\textwidth}
        \centering
        \includegraphics[width=\linewidth]{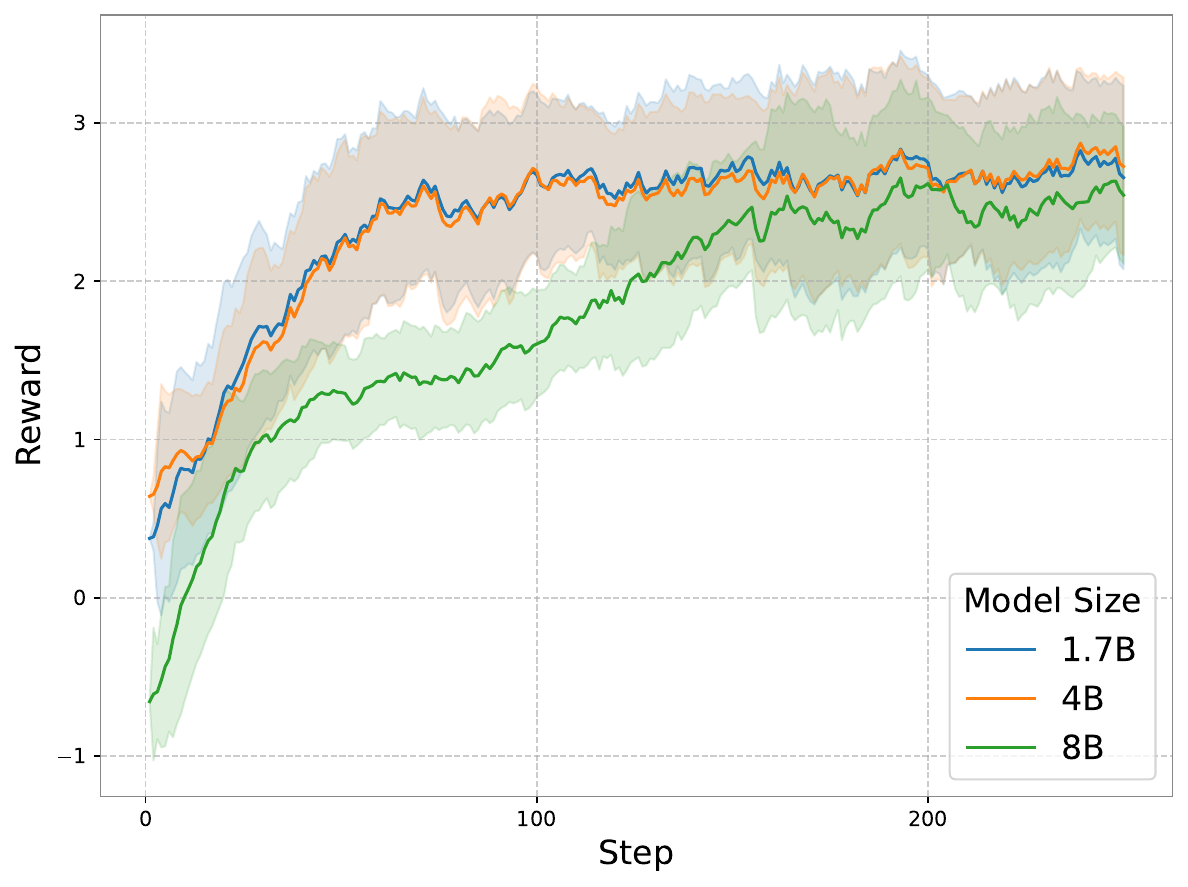}
        \caption{Outcome-Only}
    \end{subfigure}\hfill
    \begin{subfigure}{0.24\textwidth}
        \centering
        \includegraphics[width=\linewidth]{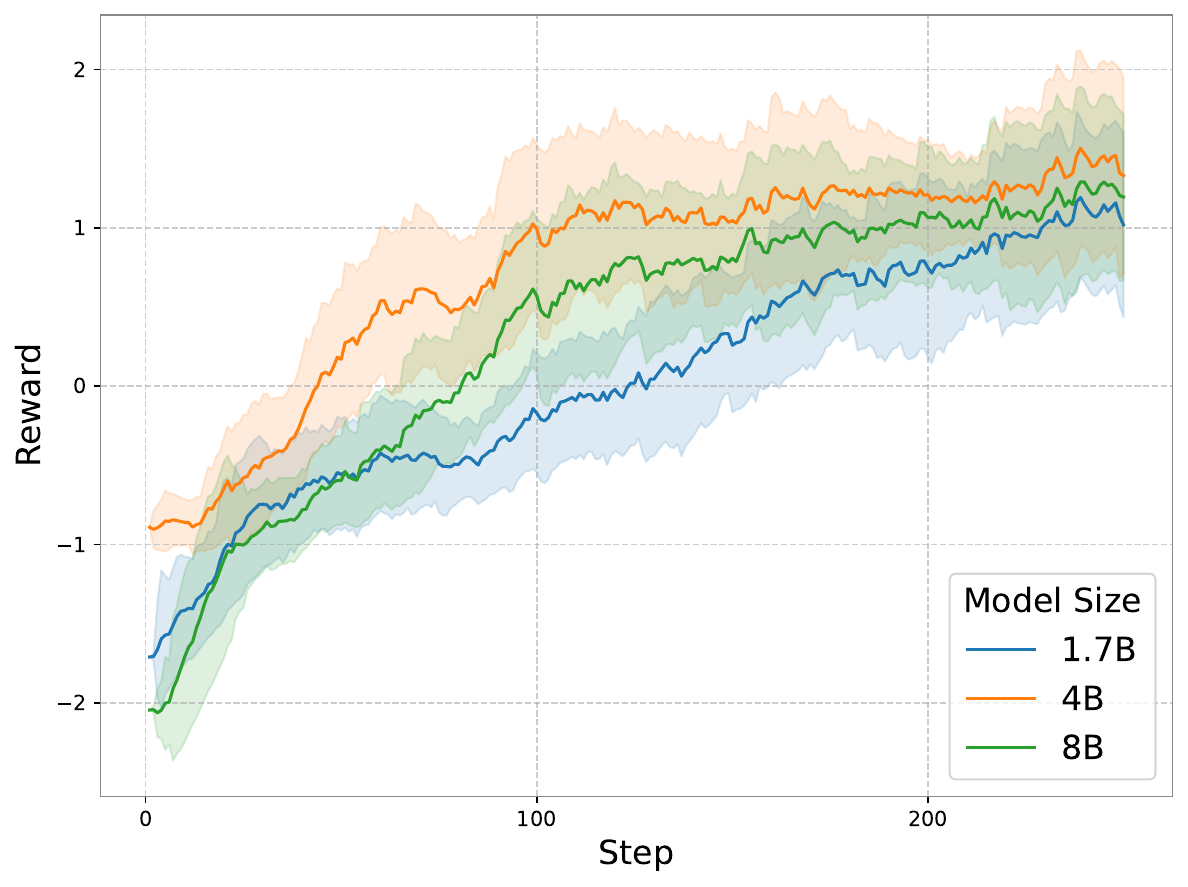}
        \caption{Hybrid (Total)}
    \end{subfigure}\hfill
    \begin{subfigure}{0.24\textwidth}
        \centering
        \includegraphics[width=\linewidth]{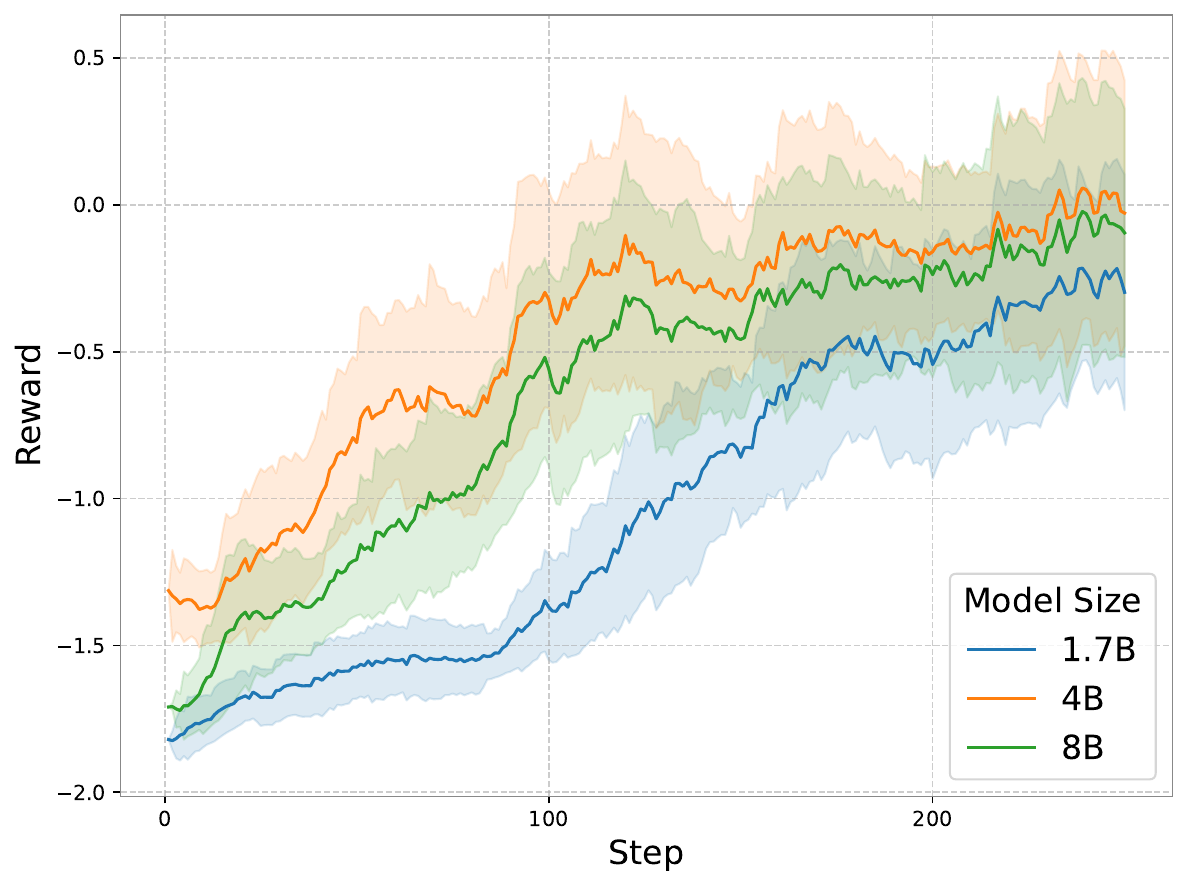}
        \caption{Hybrid (Process)}
    \end{subfigure}\hfill
    \begin{subfigure}{0.24\textwidth}
        \centering
        \includegraphics[width=\linewidth]{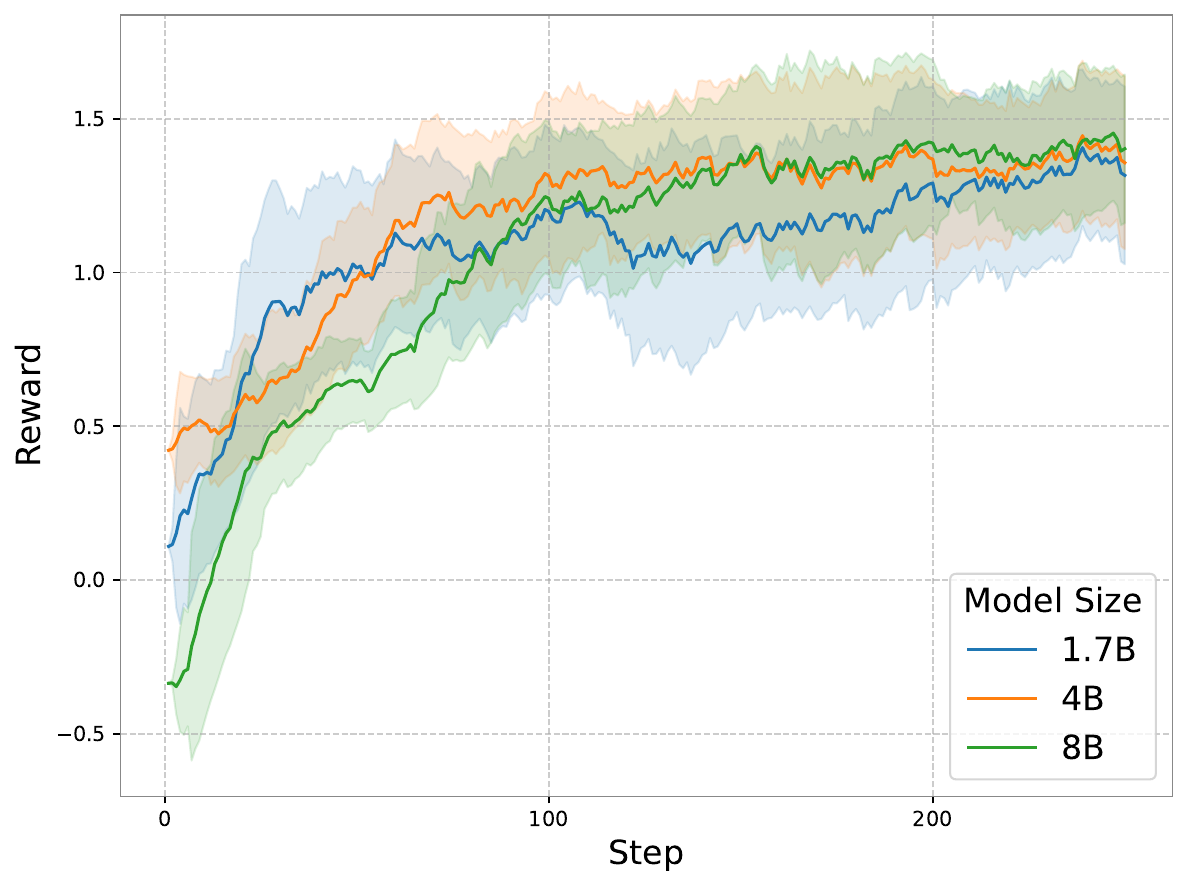}
        \caption{Hybrid (Outcome)}
    \end{subfigure}
    \caption{
        \textbf{Single-agent reward trajectories.}
        Reward progression over training steps for (a) outcome-only reward, (b) hybrid reward, (c) process component of the hybrid reward, and (d) outcome component of the hybrid reward.
    }
    \label{fig:single_agent_1x4_rewards}
\end{figure*}

%% file: figures/suppl/MAS_4B_reward_trajectories_longer.tex
\begin{figure*}[!htb]
\centering
\begin{minipage}{0.85\textwidth}
\centering
    \begin{subfigure}{0.48\textwidth}
        \centering
        \includegraphics[width=\linewidth]{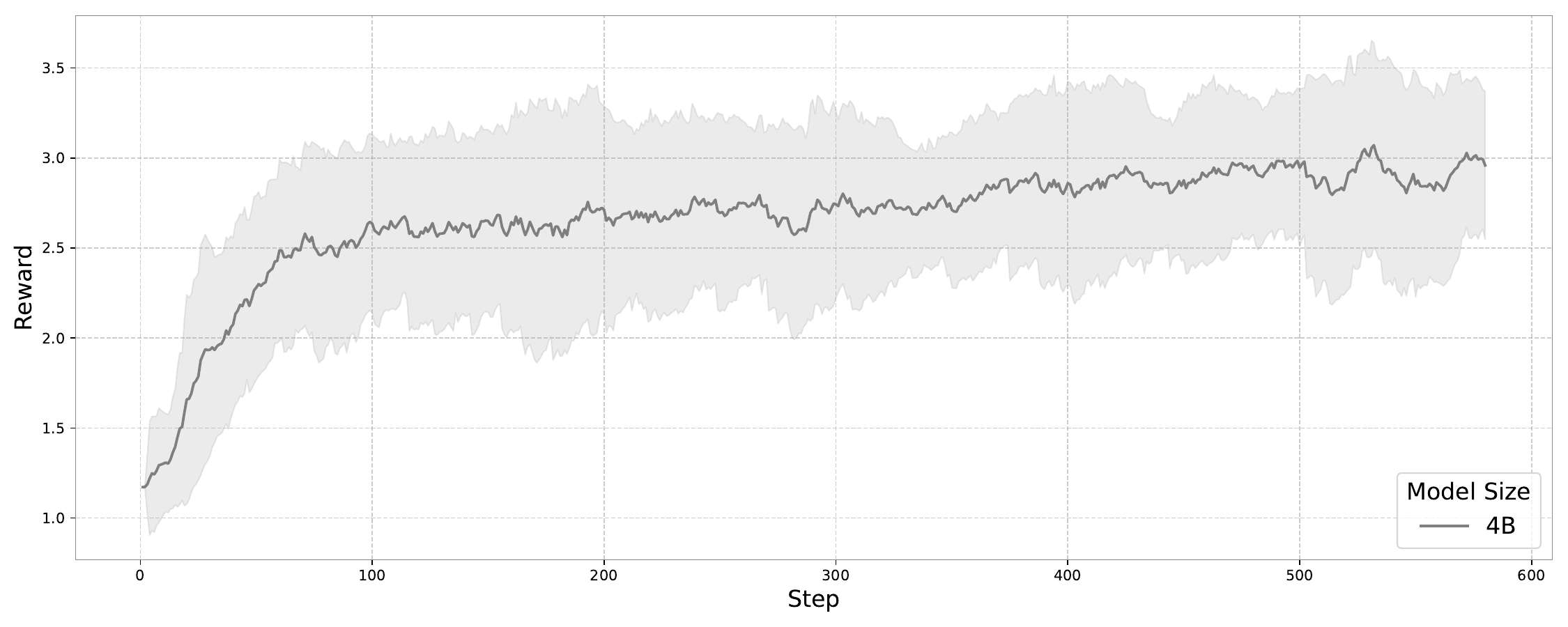}
        \caption{Outcome-Only}
    \end{subfigure}
    \hfill
    \begin{subfigure}{0.48\textwidth}
        \centering
        \includegraphics[width=\linewidth]{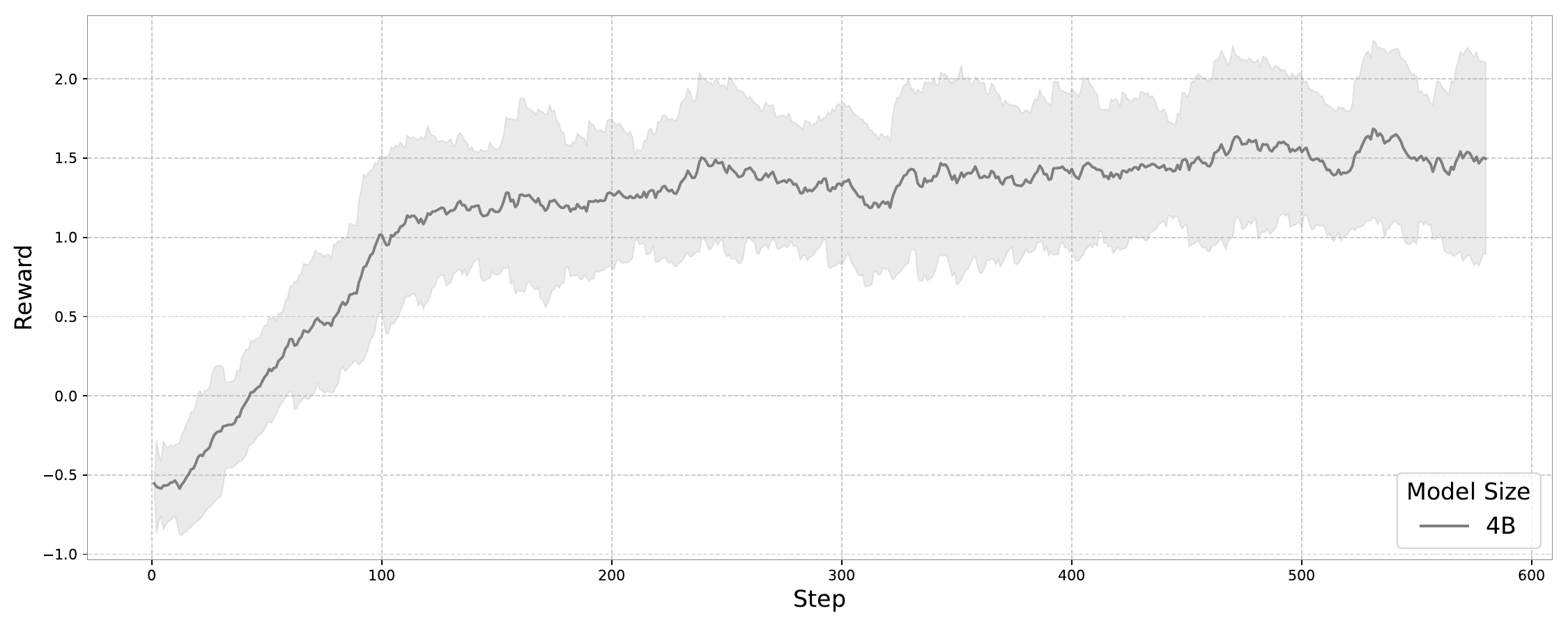}
        \caption{Hybrid (Total)}
    \end{subfigure}

    \vspace{0.5em}

    \begin{subfigure}{0.48\textwidth}
        \centering
        \includegraphics[width=\linewidth]{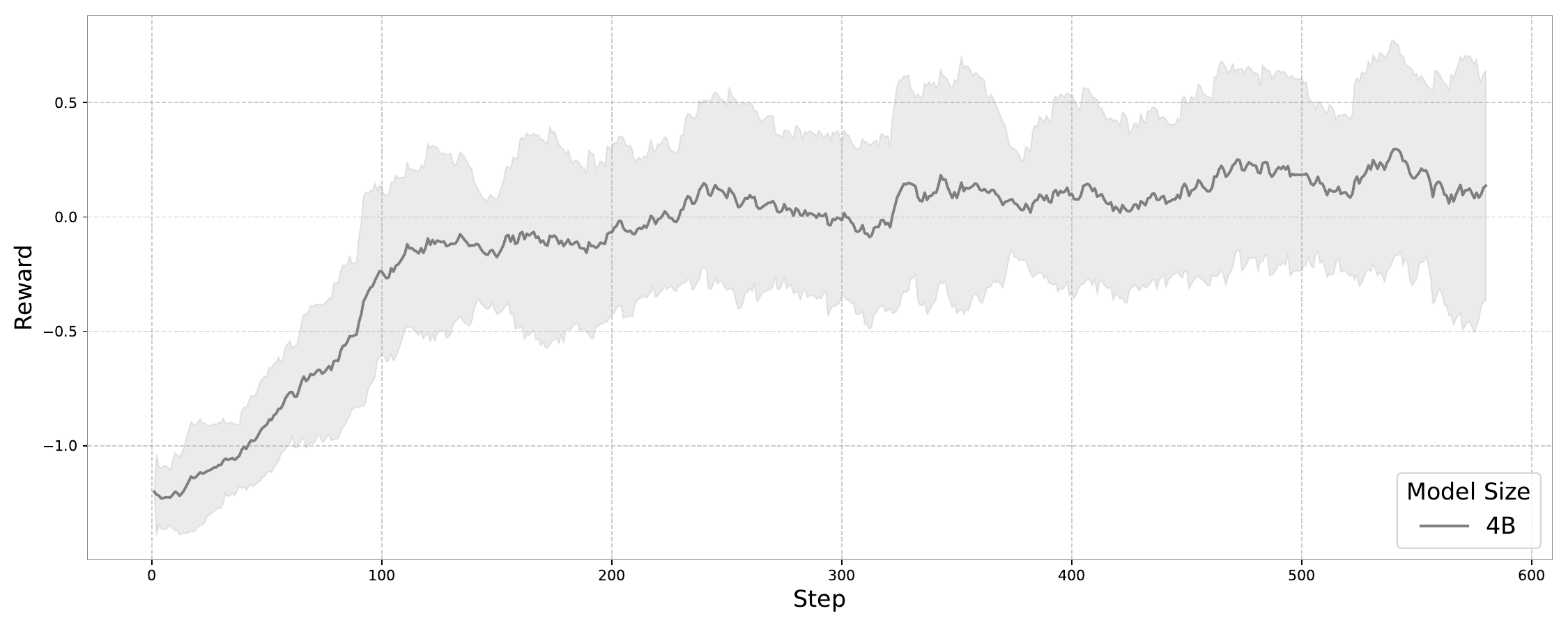}
        \caption{Hybrid (Process)}
    \end{subfigure}
    \hfill
    \begin{subfigure}{0.48\textwidth}
        \centering
        \includegraphics[width=\linewidth]{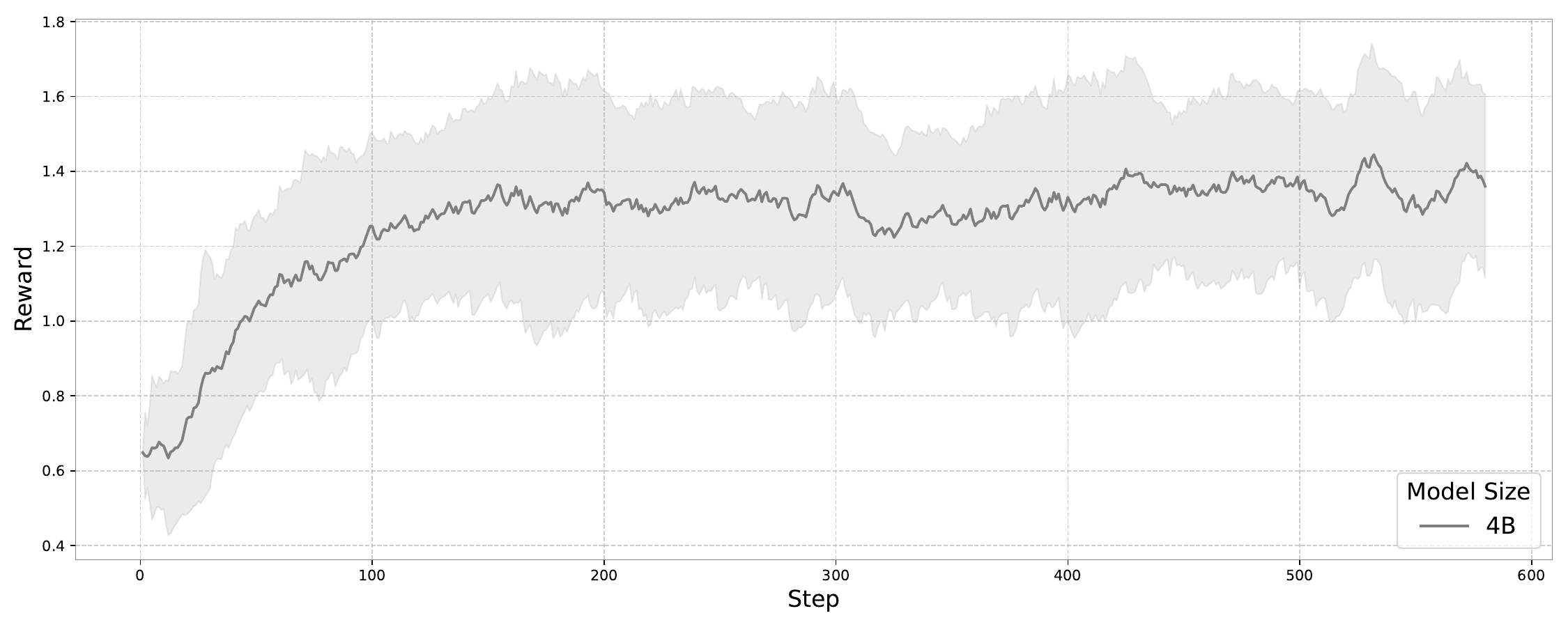}
        \caption{Hybrid (Outcome)}
    \end{subfigure}
\end{minipage}
\caption{\textbf{Qwen3-4B supervisor reward trajectories.}
Rewards over 10 training epochs under outcome-only and hybrid supervision, illustrating convergence pattern.}
\label{fig:mas_4b_appendix_2x2}
\end{figure*}

%% file: figures/suppl/cubic_f1_overlay.tex
\begin{figure}[!htb]
    \centering
    \begin{subfigure}[b]{0.48\textwidth}
        \centering
        \includegraphics[width=0.90\linewidth]{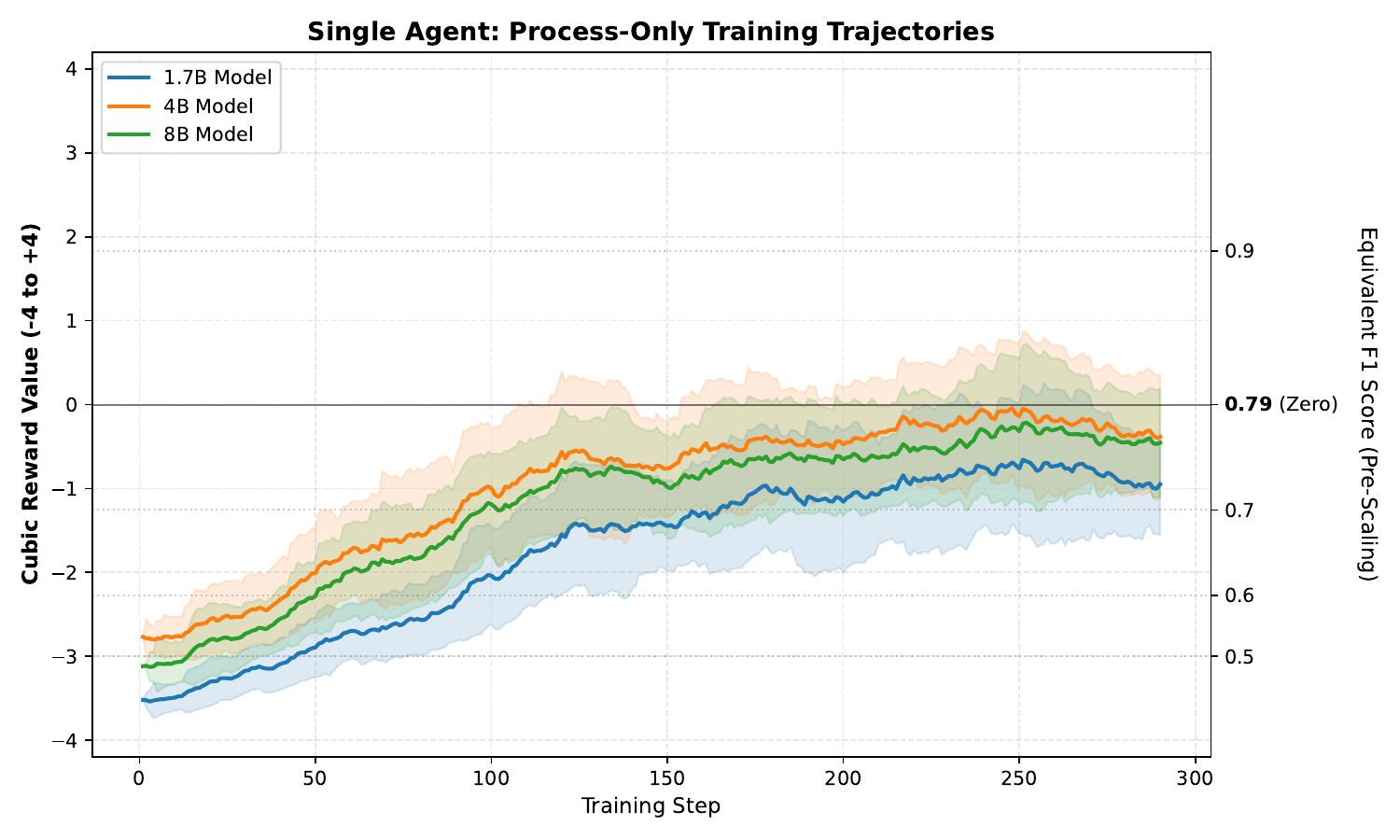}
        \caption{Single Agent}
        \label{fig:process_only_single}
    \end{subfigure}
    \hfill
    \begin{subfigure}[b]{0.48\textwidth}
        \centering
        \includegraphics[width=0.90\linewidth]{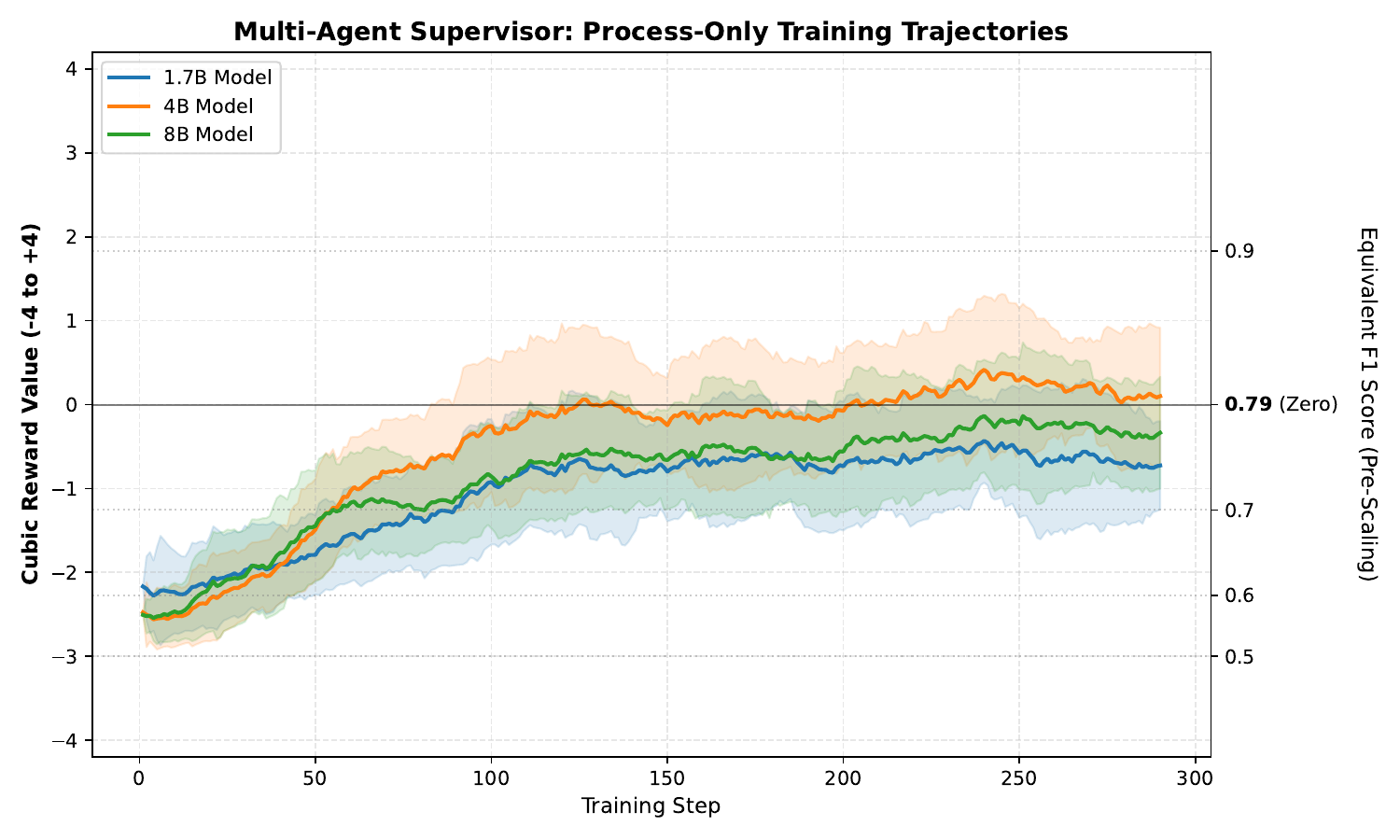}
        \caption{Multi-Agent Supervisor}
        \label{fig:process_only_mas}
    \end{subfigure}
    \caption{\textbf{Process-only training trajectories across model sizes.} The right axis shows the equivalent F1 score before cubic scaling. The horizontal line at 0.79 indicates the point where cubic-scaled process reward is at zero. For simplicity, the effect of the format penalty is not considered.}
    \label{fig:process_only_trajectories}
\end{figure}

%% file: tables/suppl/hps.tex
\begin{table}[!htb]
\caption{Configuration for GRPO training}
\label{tab:training_hparams}
\centering
\small
\begin{tabular}{ll}
\toprule
\textbf{Category} & \textbf{Value} \\
\midrule
\multicolumn{2}{l}{\textbf{Data Configuration}} \\
\midrule
Train Batch Size & 16 \\
Validation Rollouts per Prompt & 4 \\
Max Prompt Length & 8192 \\
Max Response Length & 4096 \\
Overlong Prompt Filter & True \\
Truncation Policy & error \\
\midrule
\multicolumn{2}{l}{\textbf{Optimisation (GRPO)}} \\
\midrule
Advantage Estimator & GRPO \\
Learning Rate & $1\times10^{-6}$ \\
PPO Mini Batch Size & 8 \\
Clip Ratio (low / high) & 0.2 / 0.35 \\
KL in Reward & False \\
KL Loss Used & False \\
Epochs & 5 \\
\midrule
\multicolumn{2}{l}{\textbf{Rollout Configuration}} \\
\midrule
Rollout Engine & vLLM \\
Number of Rollouts per Prompt & 8 \\
Sampling Temperature & 0.8 \\
Top-$p$ / Top-$k$ & 0.95 / 20 \\
GPU Memory Utilization & 0.4 \\
Use Inference Chat Template & False \\
\midrule
\multicolumn{2}{l}{\textbf{Multi-turn \& Tool Use}} \\
\midrule
Multi-turn Enabled & True \\
Max Assistant Turns / User Turns & 2 / 2 \\
Max Tool Response Length & 1024 (truncate: middle) \\
\midrule
\multicolumn{2}{l}{\textbf{Training \& Logging}} \\
\midrule
Save Frequency (Steps) & 50 \\
Test Frequency (Steps) & 50 \\
Validation Before Training & True \\
\bottomrule
\end{tabular}
\end{table}

%% file: tables/suppl/SFT_hps.tex
\begin{table}[!htb]
\caption{Training setup for sub-agent SFT}
\label{tab:sft_training_hparams}
\centering
\small
\begin{tabular}{ll}
\toprule
\textbf{Category} & \textbf{Value} \\
\midrule
\multicolumn{2}{l}{\textbf{Data Configuration}} \\
\midrule
Train Batch Size & 4 \\
Micro Batch Size & 1 \\
Max Sequence Length & 32,600 \\
Multi-turn & False \\
\midrule
\multicolumn{2}{l}{\textbf{Optimization}} \\
\midrule
Total Epochs & 4 \\
\midrule
\multicolumn{2}{l}{\textbf{Hardware}} \\
\midrule
GPUs & 4 \\
\bottomrule
\end{tabular}
\end{table}